\documentclass[runningheads]{llncs}

\newcommand{\PJ}[2]{\mathbb{P}_{\mathcal{#1}\mathcal{#2}}} 
  
\newcommand{\PI}[2]{\mathbb{P}_{\mathcal{#1}}\otimes\mathbb{P}_{\mathcal{#2}}}
\usepackage{microtype}
\usepackage{graphicx}
\usepackage{subfigure}
\usepackage{helvet}
\usepackage{booktabs} 
\usepackage{epsfig}
\usepackage{graphicx}
\usepackage{dsfont}
\usepackage{amssymb,amsmath}
\usepackage{url}
\graphicspath{{./images/}}

\usepackage{tabularx}
\usepackage{makecell}
\usepackage{bm}
\usepackage{multirow} 
\usepackage{rotating}
\usepackage{wrapfig}
\usepackage{subfigure}
\usepackage{thmtools}
\usepackage{thm-restate}
\usepackage{flexisym}
\usepackage{colortbl}
\usepackage{xcolor}
\usepackage{appendix}

\newcommand*{\myfont}{\fontfamily{ptm}\selectfont}
\usepackage{etoolbox}
\AtBeginEnvironment{table}{\myfont}
\AtBeginEnvironment{table*}{\myfont}

\newcommand{\STAB}[1]{\begin{tabular}{@{}c@{}}#1\end{tabular}}
\newcommand{\myrotate}[1]{\STAB{\rotatebox[origin=c]{90}{#1}}}
\newcommand{\ModelName}{\textsc{Domain2Vec}}

\newcommand{\etal}{\textit{et al}}
\newcommand{\TinyDA}{\textsc{TinyDA}}
\newcommand{\DomainBank}{\textsc{DomainBank}}

\newcommand{\BU}[1]{\color{black!40!green}\textbf{#1}}

\newcommand{\BV}[1]{\color{black!10!blue}{\em #1}}
\newcommand{\redbold}[1]{\color{red}{\textbf{#1}}}
\newcommand{\rebuttal}[1]{#1}

\usepackage{hyperref}
\hypersetup{
    colorlinks=true,
    linkcolor=blue,
    filecolor=cyan,      
    urlcolor=magenta,
}
\begin{document}

\title{Domain2Vec: Domain Embedding\\ for Unsupervised Domain Adaptation}
\titlerunning{Domain2Vec: Domain Embedding\\ for Unsupervised Domain Adaptation}
\author{Xingchao Peng\inst{*1} \and
Yichen Li\inst{*2} \and
Kate Saenko\inst{1,3}}
\authorrunning{Peng et al.}
\institute{Boston University, Boston, MA, USA \\ \and
Stanford University, Stanford, CA, USA \\
\and
 MIT-IBM Watson AI Lab, Boston, MA, USA\\
\email{\{xpeng,saenko\}@bu.edu, liyichen@stanford.edu}}

\maketitle
\renewcommand{\thefootnote}{\fnsymbol{footnote}}\footnotetext[1]{These authors contributed equally.}

\begin{abstract}
Conventional unsupervised domain adaptation (UDA) studies the knowledge transfer between a limited number of domains. This neglects the more practical scenario where data are distributed in numerous different domains in the real world. A technique to measure domain similarity is critical for domain adaptation performance. To describe and learn relations between different domains, we propose a novel \ModelName~model to provide vectorial representations of visual domains based on joint learning of feature disentanglement and Gram matrix. To evaluate the effectiveness of our \ModelName~model, we create two large-scale cross-domain benchmarks. The first one is \TinyDA, which contains 54 domains and about one million MNIST-style images. The second benchmark is \DomainBank~, which is collected from 56 existing vision datasets. We demonstrate that our embedding is capable of predicting domain similarities that match our intuition about visual relations between different domains. Extensive experiments are conducted to demonstrate the power of our new datasets in benchmarking state-of-the-art multi-source
domain adaptation methods, as well as the advantage of our proposed model. Data and code are available at \href{https://github.com/VisionLearningGroup/Domain2Vec}{\url{https://github.com/VisionLearningGroup/Domain2Vec}}
\keywords{Unsupervised Domain Adaptation, Domain Vectorization}

\end{abstract}


\section{Introduction}
\label{sec_intro}

Generalizing models learned on one visual domain to novel domains has been a major pursuit of machine learning in the quest for universal object recognition. The performance of the learned methods degrades significantly when tested on novel domains due to the presence of \textit{domain shift}~\cite{domainshift}.
 
Recently, Unsupervised Domain Adaptation (UDA) methods have been proposed to mitigate domain gap. For example, several learning-based UDA models~\cite{JAN,tzeng2014deep,long2015} incorporate Maximum Mean Discrepancy loss to minimize the domain discrepancy; other models propose different learning schema to align the marginal feature distributions of the source and target domains, including aligning second-order correlation~\cite{sun2015return,peng2017synthetic}, moment matching~\cite{zellinger2017central}, GAN-based alignment~\cite{CycleGAN2017,hoffman2017cycada,UNIT}, and adversarial domain confusion~\cite{adda,DANN,MCD_2018}. However, most of the current UDA methods consider domain adaptation between limited number of domains (usually one source domain and one target domain). In addition, the state-of-the-art UDA models mainly focus on aligning the feature distribution of the source domain with that of the target domain, and fail to consider the natural distance and relations between different domains. In the more practical scenarios where multiple domain exists and the relations between different domains are unclear, it is critical to evaluate the natural domain distances between source and target so to be able to select one or several domains from the source domain pool such that the target domain can achieve the best performance.

\begin{figure}[t]
    \centering
    \includegraphics[width=0.9\textwidth]{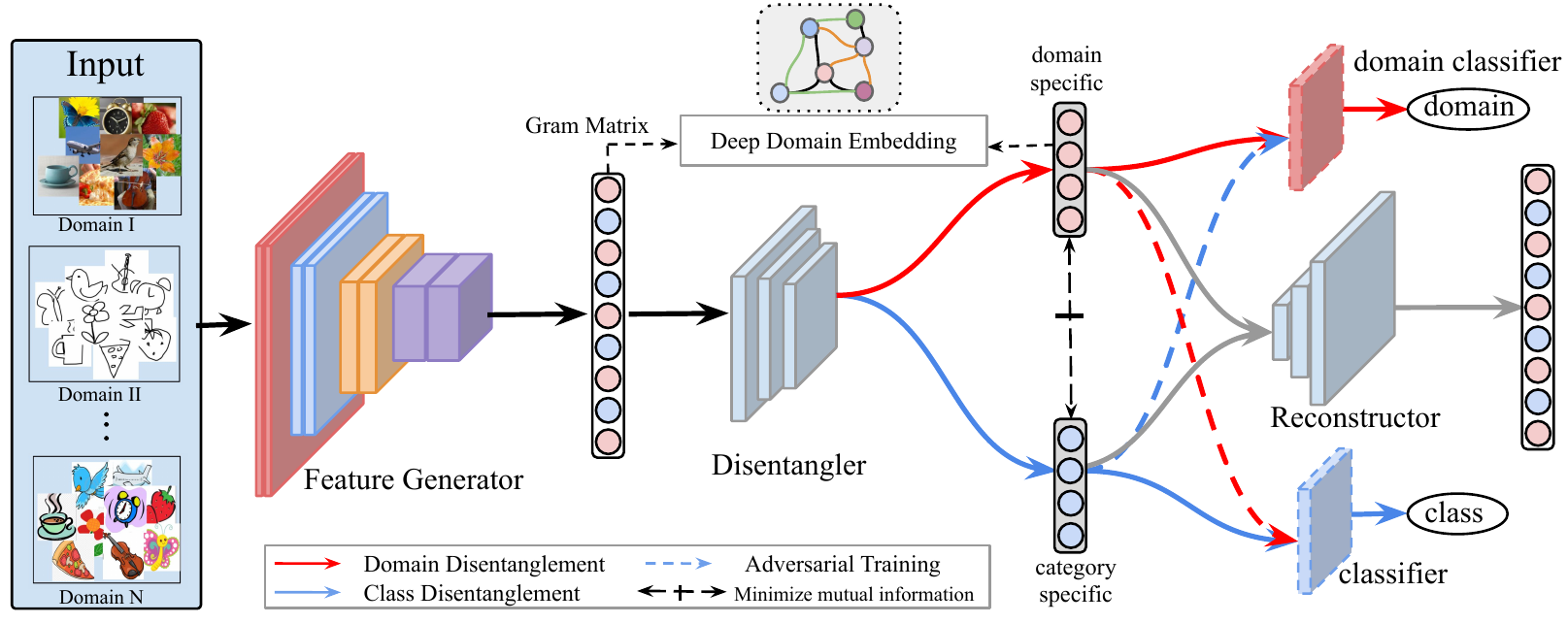}
    \caption{Our \ModelName~architecture achieve deep domain embedding by by joint learning of feature disentanglement and Gram matrix. We employ \textit{domain disentanglement} ({\color{red}red lines}) and \textit{class disentanglement} ({\color{blue}blue lines}) to extract domain-specific features and category specific features, both trained adversarially. We further apply a mutual information minimizer to enhance the disentanglement.}
    \label{figure_1}
\end{figure}

In this paper, we introduce the \ModelName~embedding to represent domains as elements of a vector space. Formally, given $N$ distinct domains  $\hat{\mathcal{D}}$ = \{$\hat{\mathcal{D}}_1, \hat{\mathcal{D}}_2, ... ,\hat{\mathcal{D}}_N\}$\footnote{In this literature, the calligraphic $\mathcal{G},\mathcal{D}$ denote Gram matrix and domains, and italic $G,D$ denote feature generator and disentangler, respectively.} domains, the aim is the learn a domain to vector mapping $\Phi:\hat{\mathcal{D}}\rightarrow V$. We would like our \ModelName~to hold the following properties: (i) given two domains $\hat{\mathcal{D}}_i$, $\hat{\mathcal{D}}_j$, the accuracy of a model trained on $\hat{\mathcal{D}}_i$ and tested on $\hat{\mathcal{D}}_j$ should be negatively correlated to the domain distance in the vector space $V$, \textit{i.e.} smaller domain distance leads to better cross-domain performance; (ii) the domain distance should match our intuition about visual relations, for example, the domain distance of two domains with building images ($\hat{\mathcal{D}}_{i}^{building}$, $\hat{\mathcal{D}}_{j}^{building}$) should be smaller than that of ($\hat{\mathcal{D}}_{i}^{building}$, $\hat{\mathcal{D}}_{j}^{car}$). Our domain embedding can be used to reason about the space of domains and solve many unsupervised domain adaptation problems. As a motivating example, we study the problem of selecting the best combination of source domains when a novel target domain emerges.

Computation of the \ModelName~embedding leverages a complementary term between the Gram matrix of deep representations and the disentangled \textit{domain-specific} feature. Gram Matrices are commonly used to build style representations that compute the correlations between different filter activations in a deep network~\cite{gatys2015neural}. Since activations of a deep network trained on a visual domain are a rich representation of the domain itself, we use Gram Matrix to capture the texture information of a domain and further obtain a stationary, multi-scale representation of the input domain. Specifically, given a domain defined by $\hat{\mathcal{D}}=\{x_j,y_j\}_{j=1}^{n_i}$ with $n_i$ ($i\in[1,N]$) examples, we feed the data through a pre-train reference convolutional neural network which we call feature generator $G$, and compute the activations of the fully connected layer as the latent representation $f_G$, as shown in Figure~\ref{figure_1}. Inspired by the feature disentanglement idea~\cite{DAL_DADA}, we introduce a disentangler $D$ to disentangle $f_G$ into \textit{domain-specific} feature $f_{ds}$ and \textit{category-specific} feature $f_{cs}$. Finally, we compute the Gram matrix of the activations of the hidden convolutional layers in the feature extractor. Given a domain $\hat{\mathcal{D}}=\{x_j,y_j\}_{j=1}^{n_i}$, we average the domain-specific features of all the training examples in $\hat{\mathcal{D}}$ as the \textit{prototype} of domain $\hat{\mathcal{D}}$. We utilize the concatenation of \textit{prototype} and the diagonal entries of the average Gram matrix as the final embedding vector of domain $\hat{\mathcal{D}}$. 
We show this embedding encodes the intrinsic properties of the domains (Sec~\ref{experiment}).

To evaluate our \ModelName~model, a large-scale benchmark with multiple domains is required. However, state-of-the-art cross-domain datasets contain only a limited number of domains. For example, the large-scale DomainNet~\cite{domainnet} that contains six domains, and the Office-31~\cite{office} benchmark that only has three domains. In this paper, we create two large-scale datasets to facilitate the research of multi-domain embedding. \TinyDA dataset is by far the largest MNIST-style cross domain dataset. It contains 54 domains and about one million training examples. Following Ganin \textit{et al}~\cite{DANN}, the images are generated by blending different foreground shapes over patches randomly cropped from the background images. The second benchmark is \DomainBank, which contains 56 domains sampled from the existing popular computer vision datasets. 

Based on \TinyDA~dataset, we validate our \ModelName~model's property on the negative correlation between the cross-domain performance and the domain distance computed by our model. Then, we show the effectiveness of our \ModelName~on multi-source domain adaptation. In addition, comprehensive experiments on \DomainBank~benchmark with openset domain adaptation and partial domain adaptation schema demonstrate that our model achieves significant improvements over the state-of-the-art methods.

The main contributions of this paper are highlighted as follows: (\textbf{i}) we propose a novel learning paradigm of deep domain embedding and develop a \ModelName~model to achieve the domain embedding; (\textbf{ii}) we collect two state-of-the-art benchmarks to facilitate research in multiple domain embedding and adaptation. (\textbf{iii}) we conduct extensive experiments on various domain adaptation settings to demonstrate the effectiveness of our proposed model.

\section{Related Work}

\textbf{Vectorial Representation Learning} Discovery of effective representations that capture salient semantics for a given task is a fundamental goal for perceptual learning. The individual dimensions in the vectorial embedding have no inherent meaning. Instead, it is the overall patterns of location and distance between vectors that machine learning takes advantage of.
GloVe~\cite{pennington2014glove} models achieve global vectorial embbedings for word by training on the nonzero elements in a word-word co-occurrence matrix, rather than on the entire sparse matrix or on individual context windows in a large corpus. DECAF~\cite{donahue2014decaf} investigates semi-supervised multi-task learning of deep convolutional representations, where representations are learned on a set of related problems but
applied to new tasks which have too few training examples to learn a full deep representation. Modern state-of-the-art deep models~\cite{alexnet,vgg,resnet,resnext,huang2017densely} learn semantic representations with supervision and are applied to various vision and language processing tasks. Another work which is very related to our work is the \textsc{Task2Vec} model~\cite{achille2019task2vec} which leverage the Fisher Information Matrix as the vectorial representation of different tasks. However, the \textsc{Task2Vec} model mainly consider the similarity between different tasks. In this work, we mainly focus on the same task and introduce a \ModelName~framework to achieve deep domain embedding for multiple domains. Specifically, \textsc{Domain2Vec} is initially proposed in the work of Deshmukh \textit{et al}~\cite{deshmukh2018domain2vec}. However, their model is designed for domain generalization. Our model is developed independently for a different purpose.

\noindent \textbf{Unsupervised Domain Adaptation} Deep neural networks have achieved remarkable success on diverse vision tasks~\cite{resnet,renNIPS15fasterrcnn,he2017mask} but at the expense of tedious labor work on labeling data. Given a large-scale unlabeled dataset, it is expensive to annotate enough training data such that we can train a deep model that generalizes well to that dataset. Unsupervised Domain Adaptation~\cite{office,long2015,DANN,MCD_2018,domainnet,DAL_DADA,SE} provides an alternative way by transferring knowledge from a different but related domain (source domain) to the domain of interest (target domain). Specifically, unsupervised domain adaptation (UDA) aims to transfer the knowledge learned from one or more labeled source domains to an unlabeled target domain. Various methods have been proposed, including discrepancy-based UDA approaches~\cite{JAN,ddc,ghifary2014domain,peng2017synthetic}, adversary-based approaches~\cite{cogan,adda,ufdn}, and reconstruction-based approaches~\cite{yi2017dualgan,CycleGAN2017,hoffman2017cycada,kim2017learning}. These models are typically designed to tackle single source to single target adaptation. Compared with single source  adaptation, multi-source domain adaptation (MSDA) assumes that training data are collected from multiple sources. Originating from the theoretical analysis in~\cite{ben2010theory,Mansour_nips2018,crammer2008learning}, MSDA has been applied to many practical applications~\cite{xu2018deep,duan2012exploiting,domainnet}. Specifically, Ben-David \etal~\cite{ben2010theory} introduce an $\mathcal{H}\Delta\mathcal{H}$-divergence between the weighted combination of source domains and a target domain. Different from the previous work, we propose a \ModelName~model to evaluate the natural distances between different domains.

\noindent \textbf{Deep Feature Disentanglement} Deep neural networks are known to extract features where multiple hidden factors are highly entangled~\cite{zhuang2015supervised}. Learning disentangled representations can help to model the relevant factors of data variation as well as evaluate the relations between different domains by extracting the domain-specific features. To this end, recent work~\cite{mathieu2016disentangling,makhzani2015adversarial,ufdn,cisac_gan} leverages generative adversarial networks (GANs)~\cite{gan} or variational autoencoders (VAEs)~\cite{vae} to learn the interpretable representations. Under the multi-domain setting, Liu \textit{et al.}~\cite{ufdn} propose a unified feature disentanglement framework to learn domain-invariant features from data across different domains. Odena \textit{et al.}~\cite{cisac_gan} introduce an auxiliary classifier GAN (AC-GAN) to achieve representation disentanglement under supervised setting. Recent work ~\cite{drit,DAL_DADA} propose to disentangle the features into a domain-invariant content space and a domain-specific attributes space, producing diverse outputs without paired training data. In this paper, we propose a cross-disentanglement schema to disentangle the deep features into \textit{domain-specific} and \textit{category-specific} features.

\section{Domain2Vec}
\label{domain2vec}
We define the domain vectorization task as follows: given N domains $\hat{\mathcal{D}}$ = \{$\hat{\mathcal{D}}_1, \hat{\mathcal{D}}_2, ... ,\hat{\mathcal{D}}_N\}$ domains, the aim is the learn a domain to vector mapping $\Phi:\hat{\mathcal{D}}\rightarrow V$, which is capable of predicting domain similarities that match our intuition about visual relations between different domains. Our \ModelName~includes two components: we first leverage feature disentanglement to generate the \textit{domain-specific} features, and then we achieve deep domain embedding by the joint learning of Gram Matrix of the latent representations and the \textit{domain-specific} features.  

\subsection{Feature Disentanglement}

Given an image-label pair (x,y), a deep neural network is a family of function $p_{\theta}(y|x)$, trained to approximate the posterior $p(y|x)$ by minimizing the cross entropy loss $H_{p_{\theta},\hat{p}}(y|x)=\mathbb{E}_{x,y}[-\log p_{\theta}(y|x)]$, where $\hat{p}$ is the empirical distribution defined by the $i$-th domain $\hat{\mathcal{D}}_i=\{x_j,y_j\}_{j=1}^{n_i}$ with $n_i$ training examples, $i \in [1,n]$. It is beneficial, especially in domain vectorization task, to think of the deep neural network as composed of two parts: a feature generator which computes the latent representations $f_{\theta}=\phi_{\theta}(x)$ of the input data, and a classifier which encodes the distribution $p(y|x)$ given the representation $f_{\theta}$.  

The latent representations $f_{\theta}=\phi_{\theta}(x)$ are highly entangled with multiple hidden factors. We propose to disentangle the hidden representations to \textit{domain-specific} and \textit{category-specific} features. Figure~\ref{figure_1} shows the proposed model. Given $N$ domains, the feature extractor $G$ maps the input data to a latent feature vector $f_G$, which contains both the domain-specific and category-specific factors. The disentangler $D$ is trained to disentangle the feature $f_G$ to \textit{domain-specific} feature $f_{ds}$ and \textit{category-specific} feature $f_{cs}$ with cross-entropy loss and adversarial training loss. The feature reconstructor $R$ is responsible to recover $f_G$ from ($f_{ds}$,$f_{cs}$) pair, aiming to keep the information integrity in the disentanglement process. To enhance the disentanglement, we follow Peng \textit{et al}~\cite{DAL_DADA} to apply a mutual information minimizer between $f_{ds}$ and $f_{cs}$. A category classifier $C$ is trained with class labels to predict the class distributions and a domain classifier $DC$ is trained with domain labels to predict the domain distributions. In addition, the cross-adversarial training step removes domain information from $f_{cs}$ and category information from $f_{ds}$. We next describe each component in detail. 

\noindent \textbf{Category Disentanglement} Given an input image $x$, the feature generator $G$ computes the latent representation $f_G$. Our category disentanglement is achieved by two-step adversarial training. First, we train the disentangler $D$ and the $k$-way category classifier $C$ to correctly predict the class labels, supervised by the cross-entropy loss:
\begin{equation}
    \mathcal{L}_{ce}^{class} = -\sum_{i=1}^{N}\mathbb{E}_{(x,y_c)\sim\hat{\mathcal{D}}_i} \sum_{k=1}^{K}\mathds{1} [k=y_c]log(C(f_{cs}))
    \label{equ_cross_entropy}
\end{equation}
where $f_{cs}=D(G(x))$ and $y_c$ indicates the class label. 

In the second step, we aim to remove the domain-specific information from $f_{cs}$. Assume that we already have a well-trained domain classifier (which is easy to with by Equation~\ref{equ_cross_entropy_domain}), we freeze the parameters in the domain classifier $DC$ and train the disentangler to generate $f_{cs}$, aiming to fool the domain classifier. This can be achieved by minimizing the negative entropy of the predicted domain distribution:
\begin{equation}
\mathcal{L}^{class}_{ent} =  -\sum_{i=1}^{N} \frac{1}{n_i} \sum_{j=1}^{n_i} \log DC(f_{cs})
\label{equ_entropy_loss}
\end{equation}
This adversarial training process corresponds to the blue dotted line in Figure~\ref{figure_1}. The above adversarial training process forces the generated \textit{category-specific} feature $f_{cs}$ contains no \textit{domain-specific} information.

\noindent \textbf{Domain Disentanglement} To achieve deep domain embedding, disentangling \textit{category-specific} features is not enough, as it fails to describe the relations between different domains. We introduce domain disentanglement to disentangle the domain-specific features from the latent representations. Previous adversarial-alignment based UDA models~\cite{adda,DAL_DADA} propose to leverage a domain classifier to classify the input feature as source or target. However, the proposed domain classifier is a binary classifier, which can not be applied to our case directly. Similar to category disentanglement, our domain disentanglement is achieved by two step adversarial training. We first train the feature generator $G$ and disentangler $D$ to extract the domain-specific feature $f_{ds}$, supervised by domain labels and cross-entropy loss:
\begin{equation}
    \mathcal{L}_{ce}^{domain} = -\mathbb{E}_{(x,y_d)\sim\hat{\mathcal{D}}} \sum_{k=1}^{N}\mathds{1} [k=y_d]log(DC(f_{ds}))
    \label{equ_cross_entropy_domain}
\end{equation}
where $f_{ds}=D(G(x))$ and $y_d$ denotes the domain label. 

In the second step, we aim to remove the category-specific information from $f_{ds}$. Assume the classifier $C$ has been well-trained in the category disentanglement, we freeze the parameters in the category classifier $C$ and train the disentangler to generate $f_{ds}$, aiming to fool the category classifier $C$. Similarly, we can minimize the negative entropy of the predicted class distribution:
\begin{equation}
\mathcal{L}^{domain}_{ent} =  -\sum_{i=1}^{N} \frac{1}{n_i} \sum_{j=1}^{n_i} \log C(f_{ds})
\label{equ_entropy_loss_domain}
\end{equation}
This adversarial training process corresponds to the red dotted line in Figure~\ref{figure_1}. If a well-trained category classifier $C$ is not able to predict the correct class labels, the category-specific information has been successfully removed from $f_{ds}$.

\noindent \textbf{Feature Reconstruction} Previous literature~\cite{DAL_DADA} has shown that the deep information could be missing in the feature disentangle process, especially when the feature disentangler $D$ is composed of several fully connected and \textsc{Relu} layers and it cannot guarantee the information integrity in the feature disentanglement process. We therefore introduce a feature reconstructor $R$ to recover the original feature $f_G$ with the disentangled \textit{domain-specific} feature and \textit{category-specific} feature. The feature reconstructor $R$ has two input and will concatenate the ($f_{ds}$,$f_{cs}$) pair to a vector in the first layer. The feature vector is feed forward to several fully connected and \textsc{Relu} layers. Denoting the reconstructed feature as $\hat{f}_G$, we can train the feature reconstruction process with the following loss:
\begin{equation} 
\label{eqn:reconstruction}
\mathcal{L}_{rec} = {\lVert \hat{f}_G - f_G \rVert}_{F}^{2} + KL(q(z|f_G)||p(z))
\end{equation}
where the first term aims at recovering the original features extracted by $G$, and the second term calculates \textit{Kullback-Leibler divergence} which penalizes the deviation of latent features from the prior distribution $p(z_c)$ (as $z \sim \mathcal{N}(0,I)$).

\noindent \textbf{Mutual Information Minimization} The mutual information is a pivotal measure of the mutual dependence between two variables. To enhance the disentanglement, we minimize the mutual information between \textit{category-specific} features and \textit{domain-specific} features. Specifically, the mutual information is defined as:
\begin{equation}
    I({f}_{ds}; {f}_{cs}) = \int_{\mathcal{P} \times \mathcal{Q}} \log{\frac{d\PJ{P}{Q}}{d\PI{P}{Q}}} d\PJ{P}{Q}
\end{equation}
where $\PJ{P}{Q}$ is the joint probability distribution of (${f}_{ds}, {f}_{cs}$), and $\mathbb{P}_{\mathcal{P}} =
\int_{\mathcal{Q}} d\PJ{P}{Q}$, $\mathbb{P}_{\mathcal{Q}} = \int_{\mathcal{Q}} d\PJ{P}{Q}$ are the marginal probability of $f_{ds}$ and $f_{cs}$, respectively. The conventional mutual information is only tractable for discrete variables, for a limited family of problems where the probability distributions are unknown~\cite{mine}. To address this issue, we follow~\cite{DAL_DADA} to adopt the Mutual Information Neural Estimator (MINE)~\cite{mine} to estimate the mutual information by leveraging a neural network $T_{\theta}$: 
$I(\mathcal{P};\mathcal{Q}) = \sup_{\theta \in \Theta} \mathbb{E}_{\mathbb{P}_{\mathcal{P}\mathcal{Q}}}[T_\theta] - \log(\mathbb{E}_{\mathbb{P}_{\mathcal{P}} \otimes \mathbb{P}_{\mathcal{Q}}}[e^{T_\theta}])$. Practically, MINE can be calculated as
$I(\mathcal{P};\mathcal{Q})=\int\int{\mathbb{P}_{\mathcal{P}\mathcal{Q}}(p,q)\text{ }T(p,q,\theta)}$  -  $\log (\int\int\mathbb{P}_{\mathcal{P}}(p)\mathbb{P}_{\mathcal{Q}}(q)e^{T(p,q,\theta)})$. To avoid computing the integrals, we leverage Monte-Carlo integration to calculate the estimation:
\begin{equation}
I(\mathcal{P},\mathcal{Q})=\frac{1}{n}\sum^{n}_{i=1}T(p,q,\theta)-\log(\frac{1}{n}\sum_{i=1}^{n}e^{T(p,q',\theta)})
\label{equ_mutual_information}
\end{equation}
\rebuttal{where $(p,q)$ are sampled from the joint distribution, $q'$ is sampled from the marginal distribution $\mathbb{P}_{\mathcal{Q}}$, $n$ is number of training examples, and $T(p,q,\theta)$ is the neural network parameteralized by $\theta$ to estimate the mutual information between $\mathcal{P}$ and $\mathcal{Q}$, we refer the reader to MINE~\cite{mine} for more details}.

\subsection{Deep Domain Embedding}
Our $\ModelName$~model to learn domain to vector mapping $\Phi: \hat{\mathcal{D}}\rightarrow V$ by joint embedding of the Gram matrix and \textit{domain-specific} features. Specifically, given a domain $\hat{\mathcal{D}}=\{x_j,y_j\}_{j=1}^{n_i}$, we compute the disentangled features for all the training examples of $\hat{\mathcal{D}}$. The \textit{prototype} of domain $\hat{\mathcal{D}}$ is defined as: $P_{\hat{\mathcal{D}}} = \frac{1}{n_i}\sum f_{ds}^{j}$, denoting the average of the \textit{domain-specific} features of the examples in $\hat{\mathcal{D}}$. In addition, we compute the Gram matrix of the activations of the hidden convolutional layers in the feature extractor $G$. The Gram matrix build a style representation that computes the correlations between different filter responses. The feature correlations are given by the Gram matrix $\mathcal{G}\in\mathcal{R}^{n\times n}$, where $\mathcal{G}_{ij}$ is the inner product between the vectorised feature map between $i$ and $j$:
\begin{equation}
    \mathcal{G}_{ij}=\sum_{k}F_{ik}F_{jk}
\end{equation}
where $F$ is the vectorised feature map of the hidden convolutional layers. Since the full Gram matrix is unmanageably large for the feature extractor based on deep neural networks, we make an approximation by only leveraging the entries in the subdiagonal, main diagonal, and superdiagonal of the Gram matrix $\mathcal{G}$. We utilize the concatenation of the \textit{prototype} $P_{\hat{\mathcal{D}}}$ and the diagonals of the $\mathcal{G}$ as the final embedding of domain $\hat{\mathcal{D}}$.

\noindent \textbf{Eliminating Sparsity} The \textit{domain-specific} feature and the Gram matrix are high sparsity data, which hampers the effectiveness of our \ModelName~model. To address this issue, we leverage dimensionality reduction technique to decrease the dimensionality. Empirically, we start by using PCA to reduce the dimenionality of the data to a specific length. Then we leverage Stochastic Neighbor Embedding~\cite{tsne} to reduce the dimensionality to our desired one.

\noindent \textbf{Optimization} Our model is trained in an end-to-end fashion. We train the feature extractor $G$, category and domain disentanglement component $D$, MINE and the reconstructor $R$ iteratively with Stochastic Gradient Descent~\cite{SGD} or Adam~\cite{Adam} optimizer. The overall optimization objective is:
\begin{equation}
    \mathcal{L} = w_1\mathcal{L}^{class}+w_2\mathcal{L}^{domain} + w_3\mathcal{L}_{rec}+w_4\mathcal{I}(f_{ds},f_{cs})
\end{equation}
where $w_1,w_2,w_3,w_4$ are the hyper-parameters, $\mathcal{L}^{class}=\mathcal{L}_{ce}^{class}+\alpha\mathcal{L}_{ent}^{class}$, $\mathcal{L}^{domain}=\mathcal{L}_{ce}^{domain}+\alpha\mathcal{L}_{ent}^{domain}$ denote the category disentanglement loss and domain disentanglement loss.

\section{Experiments}
\label{experiment}

We test \ModelName~on two large-scale datasets we created. Our experiments aim to test both qualitative properties of the domain embedding and its performance on multi-source domain adaptation, openset domain adaptation and partial domain adaptation. In the main paper, we only report major results; more implementation details are provided in the supplementary material. Our \ModelName~is implemented in the PyTorch platform. In the main paper, we only show the main experimental results, detailed experimental settings can be seen in the supplementary material.

\subsection{Dataset}
To evaluate the domain-to-vector mapping ability of our \ModelName~model, a large-scale dataset with multiple domains is desired. Unfortunately, existing UDA benchmarks~\cite{office,officehome,domainnet,peng2017visda} only contain limited number of domains. These datasets provide limit benchmarking ability for our \ModelName~model. To address this issue, we collect two datasets for multiple domain embedding and adaptation, \textit{i.e.}, \TinyDA~ and \DomainBank.

\noindent \textbf{TinyDA} We create our by far the largest MNIST-style cross domain dataset to data, \TinyDA. This dataset contains 54 domains and about one million MNIST-style training examples. We generate our \TinyDA~dataset by blending different foreground shapes over patches randomly extracted from background images. This operation is formally defined for two images $I^1$, $I^2$ as $I^{\text{out}}_{ijk}=\|I^1_{ijk}-I^2_{ijk}\|$, where $i,j$ are the coordinates of a pixel and $k$ is the channel index. 
The foreground shapes are from MNIST~\cite{mnist}, USPS~\cite{usps}, EMNIST~\cite{emnist}, KMNIST~\cite{kmnist}, QMNIST~\cite{qmnist}, and  FashionMNIST~\cite{fashionmnist}. Specifically, the MNIST, USPS, QMNIST contains digit images; EMNIST dataset includes images of MNIST-style English characters; KMNIST dataset is composed of images of Japanese characters; FashionMNIST dataset contains MNIST-style images about fashion. The background images are randomly cropped from CIFAR10~\cite{cifar10} or BSDS500~\cite{bsds500} dataset. We perform three different post-processes to our rendered images: (1) replace the background with black patch, (2) replace the background with white patch, (3) convert the images to grayscale. The three post-processes, together with the original foreground images and the generated color images, form a dataset with five different modes, \textit{i.e.} White Background ({\BV{WB}}), Black Background ({\BV{BB}}), GrayScale image ({\BV{GS}}), Color ({\BV{Cr}}) image, Original image({\BV{Or}}).

\noindent \textbf{\DomainBank}\footnote{In this dataset, the \textit{domain} is defined by datasets. The data from different genres or times typically have different underlying distributions.} To evaluate our \ModelName~model on state-of-the-art computer vision datasets, we collect a large-scale benchmark, named~\DomainBank. The images of~\DomainBank~ dataset are sampled from 56 existing popular computer vision datasets such as COCO~\cite{mscoco}, CALTECH-256~\cite{griffin2007caltech}, PASCAL~\cite{pascal}, VisDA~\cite{peng2017visda}, DomainNet~\cite{domainnet}, \textit{etc}. We choose the dataset with different image modalities, illuminations, camera perspectives \textit{etc.} to increase the diversity of the domains. In total, we collect 339,772 images with image-level and domain-level annotations. Different from \TinyDA, the categories of different domains in \DomainBank~ are not identical. This property makes \DomainBank~a good testbed for Openset Domain Adaptation~\cite{busto2017openset,busto2018open} and Partial Domain Adaptation~\cite{cao2018partial}.


\begin{table*}[t!]
{\fontfamily{ptm}\selectfont 
\centering
\tiny
\setlength{\tabcolsep}{1.5pt}

\begin{minipage}{0.33\linewidth}
\tiny
\setlength{\tabcolsep}{0.6pt}
\begin{tabular}{|c|c|c|ccccc|cccc|}

\hline

\multicolumn{3}{|c|}{}                                  & \multicolumn{9}{c|}{KMNIST}                                                                                                                                                                                                                                                          \\ \cline{4-12} 
\multicolumn{3}{|c|}{}                                  & \multicolumn{5}{c|}{BSDS}                                                                                                                                & \multicolumn{4}{c|}{CIFAR}                                                                                                \\ \cline{4-12} 
\multicolumn{3}{|c|}{\multirow{-3}{*}{}}                & {\color[HTML]{3531FF} WB}                            & {\color[HTML]{3531FF} BB}                            &{\color[HTML]{3531FF} Or}                            & {\color[HTML]{3531FF} Cr}                            & {\color[HTML]{3531FF} GS}                            & {\color[HTML]{3531FF} WB}                            & {\color[HTML]{3531FF} BB}                            & {\color[HTML]{3531FF} Cr}                            & {\color[HTML]{3531FF} GS}                            \\ \hline
                         &                         & {\color[HTML]{3531FF} WB}  & \cellcolor[HTML]{E0E0E0}89.8 & 13.3                         & 12.4                         & 16.8                         & 16.4                         & \cellcolor[HTML]{FFFFFF}88.0 & 12.8                         & 14.9                         & 14.6                         \\ 
                         &                         & {\color[HTML]{3531FF} BB}  & 12.5                         & \cellcolor[HTML]{E0E0E0}94.1 & 94.3                         & 32.9                         & 30.4                         & 11.5                         & \cellcolor[HTML]{FFFFFF}92.6 & 23.3                         & 22.2                         \\ 
                         &                         &{\color[HTML]{3531FF} Or}  & 8.4                          & 56.9                         & \cellcolor[HTML]{E0E0E0}95.4 & 35.2                         & 32.9                         & 9.3                          & 62.6                         & 24.7                         & 23.2                         \\ 
                         &                         & {\color[HTML]{3531FF} Cr}  & 73.4                         & 68.6                         & 89.8                         & \cellcolor[HTML]{E0E0E0}84.2 & 69.1                         & 66.2                         & 66.5                         & 70.9                         & 56.5                         \\ 
                         & \multirow{-5}{*}{\STAB{\rotatebox[origin=c]{90}{BSDS}}}  & {\color[HTML]{3531FF} GS}  & 72.7                         & 64.0                         & 87.9                         & 67.4                         & \cellcolor[HTML]{E0E0E0}74.1 & 68.7                         & 66.7                         & 55.1                         & 59.4                         \\ \cline{2-12} 
                         &                         & {\color[HTML]{3531FF} WB}  & 83.8                         & 17.0                         & 16.2                         & 18.6                         & 18.9                         & \cellcolor[HTML]{E0E0E0}81.2 & 15.1                         & 18.8                         & 18.0                         \\ 
                         &                         & {\color[HTML]{3531FF} BB}  & 13.1                         & 90.0                         & 91.2                         & 26.0                          & 24.1                         & 11.8                         & \cellcolor[HTML]{E0E0E0}88.8 & 18.8                         & 17.9                         \\ 
                         &                         & {\color[HTML]{3531FF} Cr}  & 66.5                         & 65.8                         & 85.3                         & 81.4                         & 68.8                         & 61.6                         & 65.7                         & \cellcolor[HTML]{E0E0E0}76.1 & 65.7                         \\ 
\multirow{-9}{*}{\STAB{\rotatebox[origin=c]{90}{KMNIST}}} & \multirow{-4}{*}{\STAB{\rotatebox[origin=c]{90}{CIFAR}}} & {\color[HTML]{3531FF} GS}  & 64.5                         & 60.5                         & 85.8                         & 58.0                          & 70.7                         & 60.8                         & 63.4                         & 56.7                         & \cellcolor[HTML]{E0E0E0}66.8 \\ \hline
\end{tabular}
\end{minipage}
\begin{minipage}{0.33\linewidth}%
\tiny
\setlength{\tabcolsep}{0.6pt}
\begin{tabular}{|c|c|c|ccccc|cccc|}
\hline
\multicolumn{3}{|c|}{}                                  & \multicolumn{9}{c|}{EMNIST}                                                                                                                                                                                                                                                          \\ \cline{4-12} 
\multicolumn{3}{|c|}{}                                  & \multicolumn{5}{c|}{BSDS}                                                                                                                                & \multicolumn{4}{c|}{CIFAR}                                                                                                \\ \cline{4-12} 
\multicolumn{3}{|c|}{\multirow{-3}{*}{}}                & {\color[HTML]{3531FF} WB}                            & {\color[HTML]{3531FF} BB}                            &{\color[HTML]{3531FF} Or}                            & {\color[HTML]{3531FF} Cr}                            & {\color[HTML]{3531FF} GS}                            & {\color[HTML]{3531FF} WB}                            & {\color[HTML]{3531FF} BB}                            & {\color[HTML]{3531FF} Cr}                            & {\color[HTML]{3531FF} GS}                            \\ \hline
                         &                         & {\color[HTML]{3531FF} WB}  & \cellcolor[HTML]{E0E0E0}86.6 & 2.9                          & 2.8                          & 8.1                          & 8.6                          & 83.2                         & 5.1                          & 6.9                          & 7.5                          \\ 
                         &                         & {\color[HTML]{3531FF} BB}  & 3.6                          & \cellcolor[HTML]{E0E0E0}87.3 & 88.0                         & 23.4                         & 18.1                         & 4.2                          & \cellcolor[HTML]{FFFFFF}82.8 & 14.9                         & 13.4                         \\ 
                         &                         &{\color[HTML]{3531FF} Or}  & 12.0                         & 31.1                         & \cellcolor[HTML]{E0E0E0}91.3 & 33.4                         & 32.2                         & 11.1                         & 33.6                         & 21.1                         & 21.2                         \\ 
                         &                         & {\color[HTML]{3531FF} Cr}  & 59.1                         & 47.0                         & 85.8                         & \cellcolor[HTML]{E0E0E0}80.0 & 60.8                         & 47.9                         & 42.0                         & 60.0                         & 42.7                         \\ 
                         & \multirow{-5}{*}{\STAB{\rotatebox[origin=c]{90}{BSDS}}}  & {\color[HTML]{3531FF} GS}  & 59.4                         & 46.7                         & 82.5                         & 56.1                         & \cellcolor[HTML]{E0E0E0}65.9 & 52.2                         & 46.8                         & 41.2                         & 44.6                         \\ \cline{2-12} 
                         &                         & {\color[HTML]{3531FF} WB}  & \cellcolor[HTML]{FFFFFF}87.8 & 13.9                         & 4.5                          & 15.3                         & 16.7                         & \cellcolor[HTML]{E0E0E0}86.1 & 12.2                         & 13.0                         & 13.6                         \\ 
                         &                         & {\color[HTML]{3531FF} BB}  & 2.1                          & 85.4                         & 87.1                         & 18.1                         & 17.1                         & 1.9                          & \cellcolor[HTML]{E0E0E0}82.7 & 12.0                         & 12.5                         \\ 
                         &                         & {\color[HTML]{3531FF} Cr}  & 58.2                         & 48.9                         & 83.5                         & 76.1                         & 59.6                         & 48.4                         & 44.7                         & \cellcolor[HTML]{E0E0E0}67.8 & 55.0                         \\ 
\multirow{-9}{*}{\STAB{\rotatebox[origin=c]{90}{EMNIST}}} & \multirow{-4}{*}{\STAB{\rotatebox[origin=c]{90}{CIFAR}}} & {\color[HTML]{3531FF} GS}  & 46.6                         & 46.5                         & 81.1                         & 48.1                         & 63.2                         & 43.8                         & 48.8                         & 45.3                         & \cellcolor[HTML]{E0E0E0}57.4 \\ \hline
\end{tabular}
\end{minipage}
\begin{minipage}{0.3\linewidth}%
\tiny
\setlength{\tabcolsep}{0.6pt}
\begin{tabular}{|c|c|c|ccccc|cccc|}
\hline
\multicolumn{3}{|c|}{}                                        & \multicolumn{9}{c|}{FashionMNIST}                                                                                                                                                                                                                                                    \\ \cline{4-12} 
\multicolumn{3}{|c|}{}                                        & \multicolumn{5}{c|}{BSDS}                                                                                                                                & \multicolumn{4}{c|}{CIFAR}                                                                                                \\ \cline{4-12} 
\multicolumn{3}{|c|}{\multirow{-3}{*}{}}                      & {\color[HTML]{3531FF} WB}                            & {\color[HTML]{3531FF} BB}                            &{\color[HTML]{3531FF} Or}                            & {\color[HTML]{3531FF} Cr}                            & {\color[HTML]{3531FF} GS}                            & {\color[HTML]{3531FF} WB}                            & {\color[HTML]{3531FF} BB}                            & {\color[HTML]{3531FF} Cr}                            & {\color[HTML]{3531FF} GS}                            \\ \hline
                               &                         & {\color[HTML]{3531FF} WB}  & \cellcolor[HTML]{E0E0E0}83.5 & 16.9                         & 29.9                         & 27.0                         & 25.6                         & 80.7                         & 16.7                         & 27.3                         & 24.9                         \\ 
                               &                         & {\color[HTML]{3531FF} BB}  & 23.6                         & \cellcolor[HTML]{E0E0E0}84.5 & 85.4                         & 38.1                         & 36.6                         & 21.1                         & 81.7                         & 28.9                         & 28.9                         \\ 
                               &                         &{\color[HTML]{3531FF} Or}  & 15.1                         & 53.6                         & \cellcolor[HTML]{E0E0E0}87.0 & 33.0                         & 33.2                         & 14.8                         & 52.2                         & 23.8                         & 25.1                         \\ 
                               &                         & {\color[HTML]{3531FF} Cr}  & 75.6                         & 68.6                         & 85.2                         & \cellcolor[HTML]{E0E0E0}81.6 & 74.4                         & 69.9                         & 54.7                         & 75.6                         & 71.3                         \\ 
                               & \multirow{-5}{*}{\myrotate{BSDS}}  & {\color[HTML]{3531FF} GS}  & 72.3                         & 66.3                         & 83.5                         & 71.5                         & \cellcolor[HTML]{E0E0E0}77.6 & 70.2                         & 61.9                         & 69.5                         & 73.2                         \\ \cline{2-12} 
                               &                         & {\color[HTML]{3531FF} WB}  & 82.9                         & 18.1                         & 27.2                         & 28.5                         & 28.6                         & \cellcolor[HTML]{E0E0E0}81.8 & 17.0                         & 29.6                         & 29.3                         \\ 
                               &                         & {\color[HTML]{3531FF} BB}  & 21.1                         & \cellcolor[HTML]{FFFFFF}84.8 & 86.2                         & 29.1                         & 28.4                         & 18.1                         & \cellcolor[HTML]{E0E0E0}82.3 & 22.1                         & 23.3                         \\ 
                               &                         & {\color[HTML]{3531FF} Cr}  & 75.1                         & 67.9                         & 85.1                         & \cellcolor[HTML]{FFFFFF}82.2 & 75.6                         & 72.4                         & 62.4                         & \cellcolor[HTML]{E0E0E0}78.6 & \cellcolor[HTML]{FFFFFF}76.6 \\ 
\multirow{-9}{*}{\myrotate{FashionMNIST}} & \multirow{-4}{*}{\myrotate{CIFAR}} & {\color[HTML]{3531FF} GS}  & 67.9                         & 61.8                         & 82.2                         & 65.2                         & 77.0                         & 66.3                         & 58.0                         & 68.7                         & \cellcolor[HTML]{E0E0E0}76.3 \\ \hline
\end{tabular}

\end{minipage}

}

\caption{Experimental results on \TinyDA. The column-wise domains are source domains, the row-wise domains are the target domain. }
\label{tiny_da_cross_domain_result}
\end{table*}

\begin{figure*}[t!]
    \begin{minipage}{\hsize}
      \centering
      \subfigure[\scriptsize t-SNE Plot ]
      {\includegraphics[width=0.31\hsize]{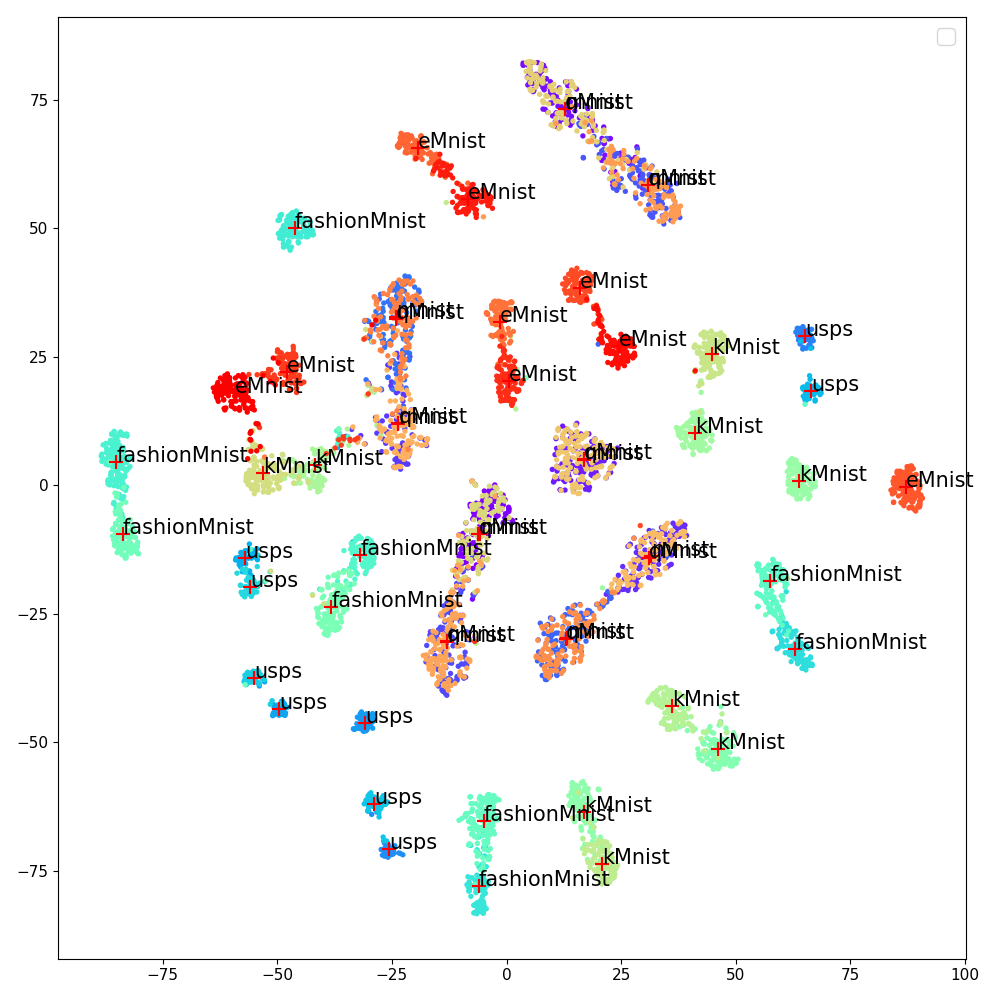}
      \label{tsne_domain_tinyDA} }
     \centering
      \subfigure[\scriptsize Domain Knowledge Graph]
      {\includegraphics[width=0.31\hsize]{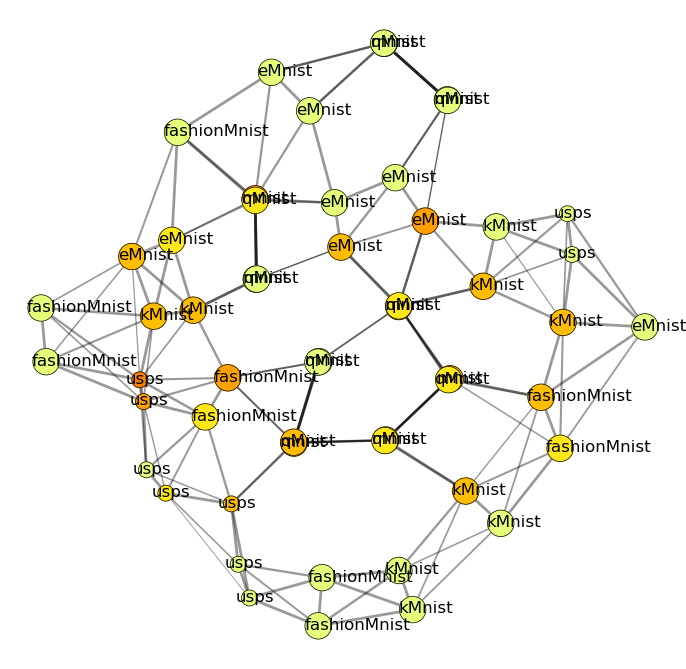}
      \label{domain_knowledge_graph}}
    \centering
      \subfigure[\scriptsize Deep Domain Embedding]
      {\includegraphics[width=0.31\hsize]{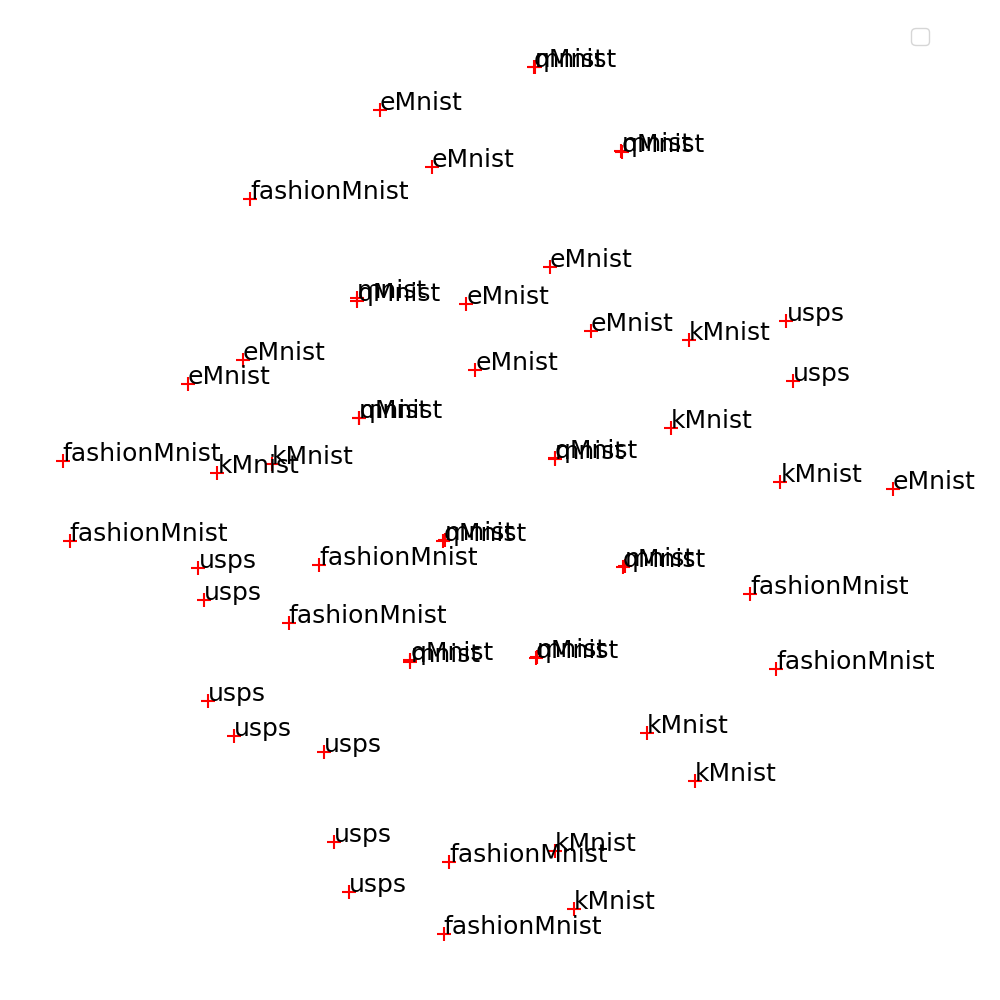}
      \label{domain_embedding} }
    \end{minipage}
  \caption{\footnotesize Deep domain embedding results of our \ModelName~model on \TinyDA~dataset: (\textbf{a}) t-SNE plot of the embedding result (color indicates different domain); (\textbf{b}) Domain knowledge graph. The size and color of the circles indicate the number of training examples and the degree of that domain, respectively. The width of the edge shows the domain distance between two domains. (\textbf{c}) The final deep domain embedding of our \ModelName~model. (Best viewed in color. Zoom in to see details.)}
  \label{domain_embedding_tinyda}
\end{figure*}

\subsection{Experiments on \TinyDA}
\noindent \textbf{Domain Embedding Results} We apply our devised \ModelName~model to \TinyDA~dataset to achieve deep domain embedding. The results are shown in Figure~\ref{domain_embedding_tinyda}. Specifically, the domain knowledge graph shows the relations between different domains in a straightforward manner. The nodes in the graph show the deep domain embedding. For each domain, we connect it with five closest domains with a edge weighted by their domain distance. The size and the color of the nodes are correlated with the number of training images in that domain and the degree of that domain, respectively. To validate that the domain distance computed with our \ModelName~is negatively correlated with the cross-domain performance, we conduct extensive experiments to calculate the cross-domain results on \TinyDA~dataset, as shown in Table~\ref{tiny_da_cross_domain_result}.
We split the cross-domain results in three sub-tables for Japanese characters (KMNIST), English characters (EMNIST) and fashion items (FashionMNIST), respectively. In each sub-table, the column-wise domains are selected as the source domain and the row-wise domains are selected as the target domain.

From the experimental results, we make three interesting observations. (\textbf{i}) For each sub-table, the performances of training and testing on the same domain (gray background) are better than cross-domain performance, except a few outliers (pink background, mainly between MNIST, USPS, and QMNIST). (\textbf{ii}) The cross-domain performance is negatively correlated with the domain distance (illustrated in Figure~\ref{domain_knowledge_graph}). We leverage Pearson correlation coefficient (PCC)~\cite{benesty2009pearson} to quantitatively demonstrate the negative correlation. The PCC can be computed as $\rho_{xy} =  \frac{\sum_{i} (x_i-\bar{x})(y_i-\bar{y})}{\sqrt{\sum_i(x_i-\bar{x})^2 }\sqrt{\sum_i(y_i-\bar{y})^2 }}$. We set the cross-domain performance and the domain distance as two variables. The PCC that we compute for our case is -0.774, which demonstrates that our \ModelName~successfully encodes the natural domain distance.

\newcommand{\NameNoSpace}{M\textsuperscript{3}SDA}
\begin{table*}[t!]
{\fontfamily{ptm}\selectfont 
\centering
\tiny
{
\begin{tabular}{l|c |c |c| c| c |c | c |c}
\multirow{1}{0.8cm}{Standards} &
\multirow{1}{0.9cm}{Models} & 
\multirow{1}{1.3cm}{\tiny{\BV{MNIST$\rightarrow$USPS}}} &  
\multirow{1}{1.4cm}{\tiny{\BV{MNIST$\rightarrow$QMNIST}}} & 
\multirow{1}{1.4cm}{\tiny{\BV{USPS$\rightarrow$MNIST}}}& 
\multirow{1}{1.4cm}{\tiny{\BV{USPS$\rightarrow$QMNIST}} }&    
\multirow{1}{1.4cm}{\tiny{\BV{QMNIST$\rightarrow$MNIST}} } & 
\multirow{1}{1.4cm}{\tiny{\BV{QMNIST$\rightarrow$USPS}} } & 
\multirow{1}{0.5cm}{{\BU{Avg}}} \\ 

 \Xhline{0.7pt} 
\multirow{8}{*}{ \begin{tabular}[c]{@{}c@{}}Single\\Best\end{tabular} } 
& Source Only & 17.7$\pm$0.21 & 83.4$\pm$0.55 & 16.4 $\pm$ 0.32 & 16.3 $\pm$ 0.25 & 83.1 $\pm$ 0.32 & 20.2$\pm$0.31 & \BU{39.5}$\pm$0.32\\

 &DAN~\cite{long2015}&  21.4$\pm$0.27&  87.1$\pm$0.64&  19.7$\pm$0.37&  19.9$\pm$0.34&  85.7$\pm$0.34&  21.8$\pm$0.37& \BU{42.6}$\pm$0.39\\
 &RTN~\cite{RTN} &  18.0$\pm$0.28&  85.0$\pm$0.58&  18.8$\pm$0.37&  20.0$\pm$0.26&  84.2$\pm$0.42&  21.3$\pm$0.34& \BU{41.2}$\pm$0.38\\
  &JAN~\cite{JAN} &   21.7$\pm$0.27&  87.6$\pm$0.64&  19.4$\pm$0.42&  18.0$\pm$0.29&  87.2$\pm$0.36&  25.1$\pm$0.33& \BU{43.2}$\pm$0.39\\
   &DANN~\cite{DANN}&  21.2$\pm$0.25&  86.1$\pm$0.55&  20.1$\pm$0.31&  19.4$\pm$0.24&  86.6$\pm$0.38&  24.0$\pm$0.34& \BU{42.9}$\pm$0.34
\\
    &ADDA~\cite{adda} &  20.3$\pm$0.31&  88.1$\pm$0.63&  18.3$\pm$0.46&  21.4$\pm$0.38&  88.5$\pm$0.39&  25.9$\pm$0.43& \BU{43.8}$\pm$0.43
\\
 &SE~\cite{SE} &  13.6$\pm$0.42&  78.1$\pm$0.87&  10.7$\pm$0.62&  11.8$\pm$0.50&  80.1$\pm$0.64&  17.0$\pm$0.55& \BU{35.2}$\pm$0.60
 \\ 
 &MCD~\cite{MCD_2018}&  23.8$\pm$0.33&  89.0$\pm$0.61&  22.3$\pm$0.36&  19.6$\pm$0.26&  86.7$\pm$0.36&  22.6$\pm$0.41& \BU{44.0}$\pm$0.39
 \\

 \Xhline{0.7pt} 
\multirow{8}{*}{ \begin{tabular}[c]{@{}c@{}}Source\\Combine\end{tabular} } 
&Source Only &  20.2$\pm$0.23&  85.7$\pm$0.59&  19.2$\pm$0.42&  20.5$\pm$0.37&  85.1$\pm$0.25&  19.2$\pm$0.40& \BU{41.6}$\pm$0.38
\\
&DAN~\cite{long2015}&  19.8$\pm$0.30&  85.4$\pm$0.64&  22.4$\pm$0.43&  21.9$\pm$0.49&  88.0$\pm$0.33&  19.2$\pm$0.48& \BU{42.8}$\pm$0.45
\\
 &RTN~\cite{RTN} &  22.9$\pm$0.27&  88.2$\pm$0.72&  19.9$\pm$0.54&  23.2$\pm$0.49&  88.1$\pm$0.29&  20.6$\pm$0.53& \BU{43.8}$\pm$0.47
\\
&JAN~\cite{JAN}&  21.8$\pm$0.29&  88.1$\pm$0.59&  22.2$\pm$0.50&  23.9$\pm$0.45&  89.5$\pm$0.36&  22.3$\pm$0.46& \BU{44.6}$\pm$0.44
 \\
&DANN~\cite{DANN}&  22.3$\pm$0.31&  87.1$\pm$0.65&  22.1$\pm$0.47&  21.0$\pm$0.46&  84.7$\pm$0.35&  19.3$\pm$0.43& \BU{42.8}$\pm$0.45
  \\
&ADDA~\cite{adda}&  25.2$\pm$0.24&  87.9$\pm$0.61&  20.5$\pm$0.46&  22.0$\pm$0.36&  88.1$\pm$0.25&  20.7$\pm$0.49& \BU{44.1}$\pm$0.40
 \\

&SE~\cite{SE}&  19.4$\pm$0.28&  82.8$\pm$0.68&  19.3$\pm$0.45&  19.3$\pm$0.45&  84.3$\pm$0.34&  18.9$\pm$0.48& \BU{40.7}$\pm$0.45
  \\

&MCD~\cite{MCD_2018}&  23.20$\pm$0.3&  91.2$\pm$0.68&  21.6$\pm$0.46&  25.8$\pm$0.37&  86.9$\pm$0.33&  23.0$\pm$0.42& \BU{45.3}$\pm$0.43
  \\

\Xhline{0.7pt}
\multirow{4}{*}{ \begin{tabular}[c]{@{}c@{}}Multi-\\Source\end{tabular} } 
&\NameNoSpace~\cite{domainnet}&  25.5$\pm$0.26&  91.6$\pm$0.63&  22.2$\pm$0.43&  25.8$\pm$0.43&  90.7$\pm$0.30&  24.8$\pm$0.41& \BU{46.8}$\pm$0.41
\\
&DCTN~\cite{xu2018deep} &  25.5$\pm$0.28&  93.10$\pm$0.7&  22.9$\pm$0.41&  \redbold{29.5$\pm$0.47}&  91.2$\pm$0.29&  26.5$\pm$0.48& \BU{48.1}$\pm$0.44
 \\

& \textbf{Domain2Vec-$\alpha$}&   27.8$\pm$0.27&  94.3$\pm$0.64&  24.3$\pm$0.52&  27.1$\pm$0.39&  89.2$\pm$0.26&  \redbold{28.1$\pm$0.41}& \BU{48.5}$\pm$0.42
\\

& \textbf{Domain2Vec-$\beta$}& \redbold{ 28.2$\pm$0.31}&  \redbold{94.5$\pm$0.63}&  \redbold{27.6$\pm$0.41}&  29.3$\pm$0.39&  \redbold{91.5$\pm$0.26}&  27.2$\pm$0.42& \redbold{49.7$\pm$0.40}

 \\



                                  
\end{tabular}
} }
\caption{
\textbf{MSDA results on the \TinyDA~ dataset.} Our model \ModelName$-\alpha$ and \ModelName$-\beta$ achieves \textbf{48.5\%} and \textbf{49.7\%} accuracy, outperforming baselines. )
}
\label{table_msda_tinyda}
\end{table*}

\noindent \textbf{Multi-Source Domain Adaptation On \TinyDA} Our \TinyDA~dataset contains 54 domains. In our experiments, we consider the MSDA between digit datasets, \textit{i.e.} MNIST, USPS, and QMNIST dataset, resulting in six MSDA settings. We choose the ``grayscale'' ({\BV{GS}}) with CIFAR10 background as the target domain. For the source domains, we remove the two ``grayscale'' domains and leverage the remaining seven domains as the source domain.

State-of-the-art multi-source domain adaptation algorithms tackle MSDA task by adversarial alignment~\cite{xu2018deep} or matching the moments of the domains~\cite{domainnet}. However, these models neglect the effect of domain distance. We incorporate our \ModelName~model to the previous work~\cite{xu2018deep,domainnet}, and devise two models, \ModelName-$\alpha$ and \ModelName-$\beta$. Specifically, the \ModelName-$\alpha$ borrows the moment matching~\cite{domainnet} idea and the training loss is weighted by the domain distance computed by our model. The \ModelName-$\beta$ is inspired by the adversarial learning~\cite{xu2018deep} and weights computed by our model is applied. Inspired by Xu \textit{et al}~\cite{xu2018deep}, we compare MSDA results with two other evaluation standards: (\textbf{i}) \textit{single best}, reporting the single best-preforming source transfer result on the test set, and (\textbf{ii}) \textit{source combine}, combining the source domains to a single domain and performing traditional single-source single target adaptation. The high-level motivations of these two evaluation schema are: the first standard evaluates whether MSDA can boost the best single source UDA results; the second standard testify whether MSDA can outperform the trivial baseline which combines the multiple source domains as a single domain.

For both \textit{single best} and \textit{source combine} experiment setting, we take the following methods as our baselines: Deep Alignment Network (\textbf{DAN})~\cite{long2015}, Joint Adaptation Network (\textbf{JAN})~\cite{JAN}, Domain Adversarial Neural Network (\textbf{DANN})~\cite{DANN}, Residual Transfer Network (\textbf{RTN})~\cite{RTN}, Adversarial Deep Domain Adaptation (\textbf{ADDA})~\cite{adda}, Maximum Classifier Discrepancy (\textbf{MCD})~\cite{MCD_2018}, and Self-Ensembling  (\textbf{SE})~\cite{SE}. For multi-source domain adaptation, we take Deep Cocktail Network (\textbf{DCTN})~\cite{xu2018deep} and Moment Matching for Multi-source Domain Adaptation (\textbf{\text{M$^3$SDA}})~\cite{domainnet} as our baselines. 

The experimental results are shown in Table~\ref{table_msda_tinyda}. The \ModelName-$\alpha$ and \ModelName-$\beta$ achieve an \textbf{48.5\%} and \textbf{49.7\%} average accuracy, outperforming other baselines. The results demonstrate that our models outperform the \textit{single best} UDA results, the \textit{combine source} results, and can boost the multi-source baselines. We argue that the performance improvement is due to the good domain embedding of our \ModelName~model.

\subsection{Experiments on DomainBank}
\label{subsect_openset}
\begin{figure*}[t]
    \begin{minipage}{\hsize}
      \centering
      \subfigure[\scriptsize t-SNE plot]
      {\includegraphics[width=0.31\hsize]{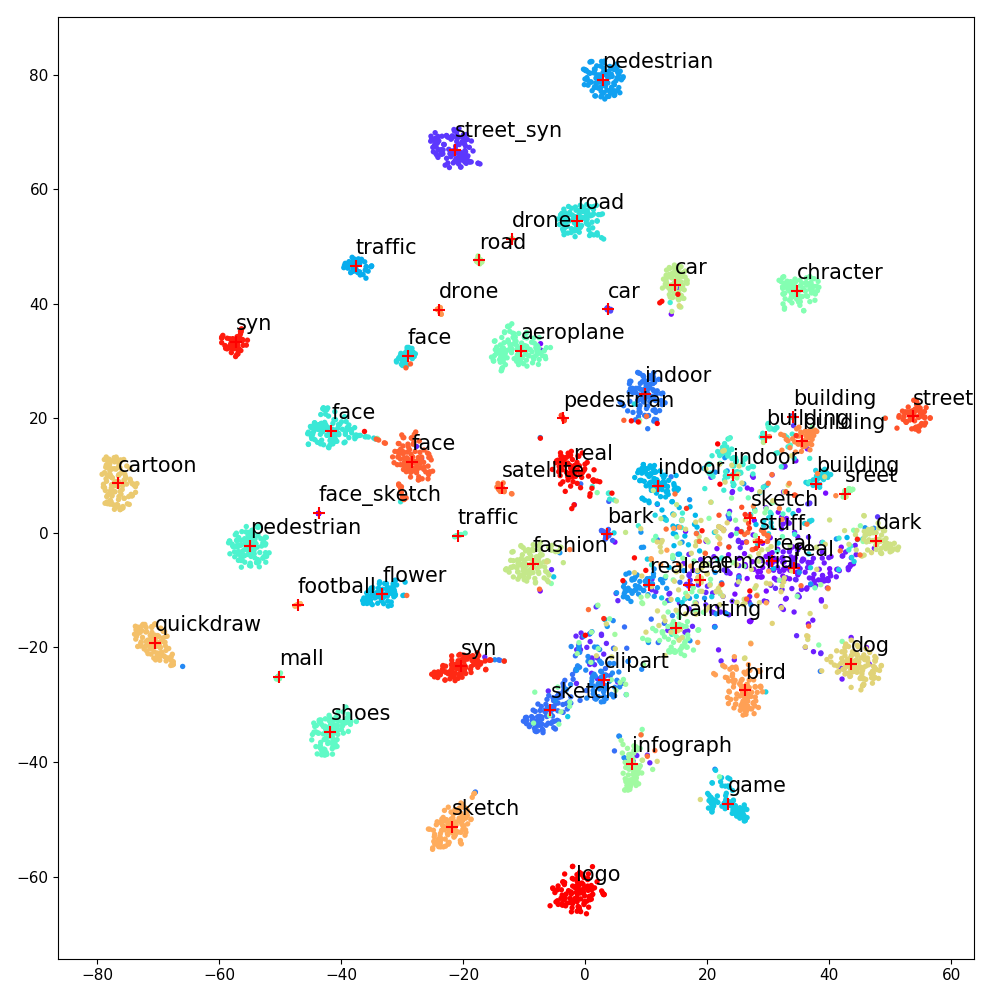}
      \label{tsne_domain_domainbank} }
     \centering
      \subfigure[\scriptsize Domain Knowledge Graph]
      {\includegraphics[width=0.31\hsize]{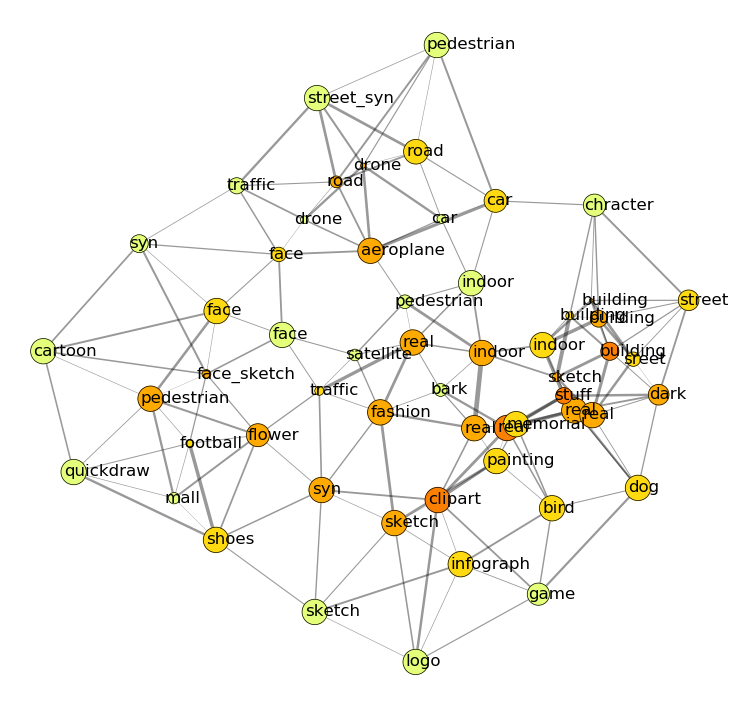}
      \label{domain_knowledge_graph_domainbank}}
    \centering
      \subfigure[\scriptsize Deep Domain Embedding]
      {\includegraphics[width=0.31\hsize]{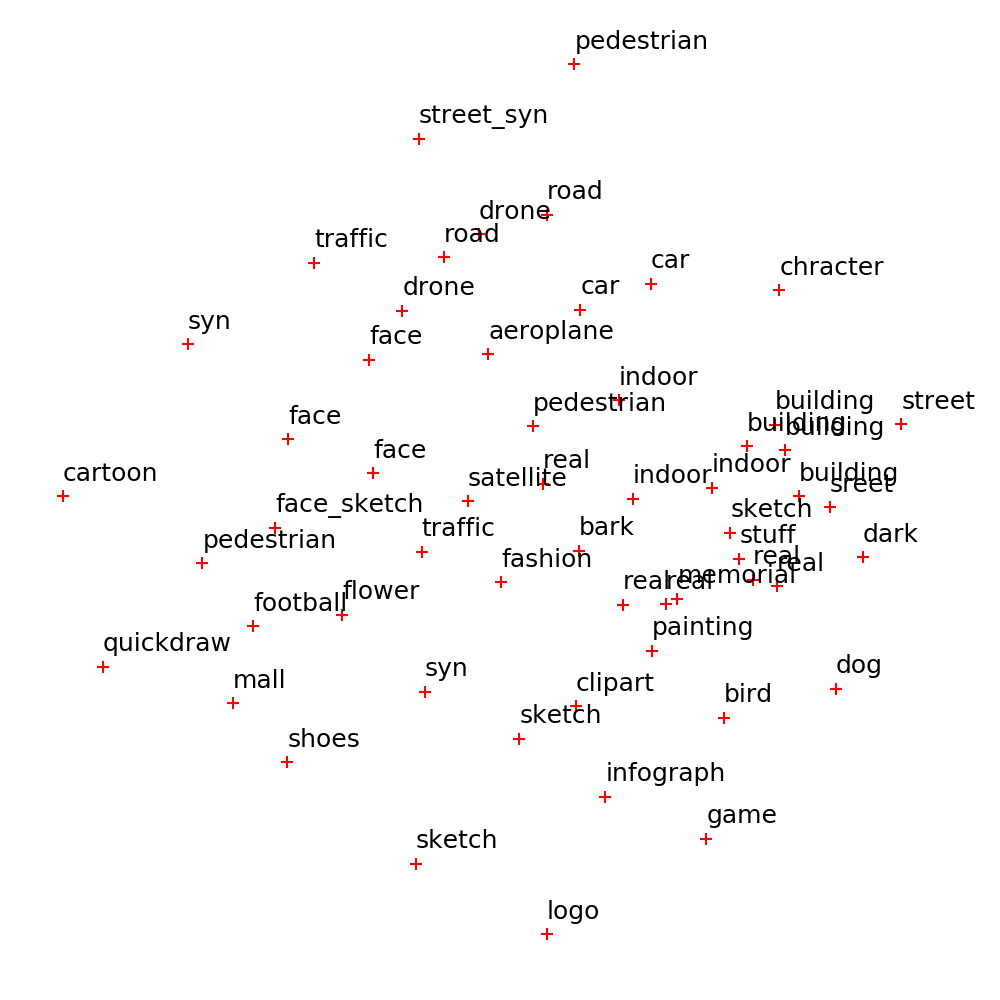}
      \label{domain_embedding_domainbank} }
    \end{minipage}
  \caption{Domain embedding results of our \ModelName~model on \DomainBank~ dataset.}
  \label{domain_embedding_domain_bank}
\end{figure*}

\noindent \textbf{Domain Embedding Results} Similar to the experiments for \TinyDA~dataset, we apply our devised~\ModelName~model to \DomainBank~dataset. The results are shown in Figure~\ref{domain_embedding_domain_bank}. Since our \DomainBank~dataset is collected from multiple existing computer vision dataset, the categories of different domains in \DomainBank~ are not identical. It is not feasible to compute the cross-domain performance directly like~\TinyDA. However, we can still make the following interesting observations: (\textbf{i}) Domains with similar contents tend to form a cluster. For example, the domains containing buildings ($\hat{\mathcal{D}}^{building}$) are close to each other in terms of the domain distance. The domains containing faces share the same property. (\textbf{ii}) The domains which contains artistic images are scattered in the exterior side of the embedding and are distinct from the domains which contains images in the wild. For example, the ``cartoon'',``syn'',``quickdraw'',``sketch'',``logo'' domains are distributed in the exterior side of the embedding space. These observations demonstrate that our \ModelName~model is capable of encoding the natural domain distance. 

\begin{table*}[t]

    \addtolength{\tabcolsep}{1.3pt}

    \centering
    
    \begin{tabular}{cccccc}
        \Xhline{0.7pt}
        Target & VisDA &  Ytb BBox & PASCAL & COCO  & Average \\
        \hline
        Source Only & 53.4$\pm0.4$ & 67.2$\pm$0.4 & 74.8 $\pm$0.4 & 80.4$\pm$0.3 &68.9\\

         Openset SVM~\cite{jain2014multi}&  53.9$\pm$0.5&  68.6$\pm$0.4&  77.7$\pm$0.4&  82.1$\pm$0.4& {70.6}
\\
        AutoDIAL &  54.2$\pm$0.5&  68.1$\pm$0.5&  75.9$\pm$0.4&  83.4$\pm$0.4& {70.4}
\\
        AODA~\cite{aoda2018saito}&  56.4$\pm$0.5&  69.7$\pm$0.4&  76.7$\pm$0.4&  82.3$\pm$0.4& {71.3}
 \\ \textbf{\ModelName} &   56.6$\pm$0.4&  70.6$\pm$0.4&  81.3$\pm$0.4&  86.8$\pm$0.4& {\textbf{73.8}}\\


        \Xhline{0.7pt}
    \end{tabular}
    \caption{Openset domain adaption on the \DomainBank~dataset.} 
    \label{openset_domain_adaptation}
\end{table*}

\noindent \textbf{Openset Domain Adaptation on \DomainBank} Open-set domain adaptation (ODA) considers classification when the target domain contains categories unknown (unlabelled) in the source domain. Our \DomainBank~dataset provides a good testbed for openset domain as the categories of different domains are not identical. Since \DomainBank~ contains 56 domains, it is infeasible to explore all the (source, target) domain combinations. Instead, in our work, we demo our model on the following four transfer setting: DomainNet~\cite{domainnet}$\rightarrow$VisDA~\cite{peng2017visda}, DomainNet$\rightarrow$Youtube BBox~\cite{real2017youtube}, DomainNet$\rightarrow$PASCAL~\cite{pascal}, DomainNet$\rightarrow$COCO. Specifically, DomainNet~\cite{domainnet} contains images with six distinct modalities and are fit to be a source domain for our openset domain adaptation.

The experimental results are shown in Table~\ref{openset_domain_adaptation}. The experimental results show that our model achieves \textbf{73.8\%} accuracy, outperforming the compared baselines.

\noindent \textbf{Partial Domain Adaptation on \DomainBank} In partial domain adaptation, the source domain label space is a superset of the target domain label space. In consistent with the openset domain adaptation, we consider the following four partial domain adaptation setting: DomainNet~\cite{domainnet}$\rightarrow$VisDA~\cite{peng2017visda}, DomainNet$\rightarrow$Youtube BBox~\cite{real2017youtube}, DomainNet$\rightarrow$PASCAL~\cite{pascal}, DomainNet$\rightarrow$COCO.

The experimental results are shown in Table~\ref{partial_domain_adaptation}.  Our model achieves \textbf{65.5\%} accuracy, outperforming the compared baselines. The experimental results demonstrate that our model can boost the performance in partial domain adaptation setting. Specifically, our model utilizes the idea of PADA~\cite{cao2018partial}, which trains a partial adversarial alignment network to tackle the partial domain adaptation task. We compute the domain distance between the sub-domains in the source training data (DomainNet) and apply the domain distance as weight in the partial adversarial alignment process.

\subsection{Ablation Study} Our model is composed of multiple component. To demonstrate the effectiveness of each component, we perform the ablation study analysis. Table~\ref{table_ablation} shows the ablation results on \TinyDA~dataset. We observe that the performance drops in most of the experiments when Mutual information minimization and Gram matrix information are \textbf{not} applied. The experimental results demonstrate the effectiveness of the mutual information minimization and Gram matrix information.

\begin{table*}[t]

    \addtolength{\tabcolsep}{1.3pt}

    \centering
    
    \begin{tabular}{cccccc}
        \Xhline{0.7pt}
        Target & VisDA &  Ytb BBox & PASCAL & COCO  & Average \\
        \hline
        Source Only & 34.5$\pm0.5$ & 74.3$\pm$0.4 & 68.2 $\pm$0.3 & 76.4$\pm$0.2 & 63.3\\
         AdaBN&  35.1$\pm$0.5&  75.6$\pm$0.5&  68.2$\pm$0.4&  78.1$\pm$0.4& {64.2}
\\
        AutoDIAL~\cite{cariucci2017autodial} &  35.2$\pm$0.6&  74.0$\pm$0.4&  68.5$\pm$0.4&  77.6$\pm$0.4& {63.8}
\\
        PADA~\cite{cao2018partial}&34.2$\pm$0.6&  76.8$\pm$0.4&  69.7$\pm$0.3&  77.7$\pm$0.4& {64.6}

\\
  \textbf{\ModelName} 
  &  36.6$\pm$0.5&  76.8$\pm$0.4&  70.0$\pm$0.3&  78.8$\pm$0.4& \textbf{{65.5}}
\\

        \Xhline{0.7pt}
    \end{tabular}
    \caption{Partial domain adaption on the \DomainBank~dataset.} 
    \label{partial_domain_adaptation}
\end{table*}

\begin{table*}[t!]
\setlength{\tabcolsep}{0.185em}
\centering
\small
{
\begin{tabular}{c| c| c| c| c| c }
{target} &
 {\tiny{\BV{MNIST}$\rightarrow$\BV{USPS} }} &  
 {\tiny{\BV{MNIST}$\rightarrow$\BV{QMNIST} }} & 
  {\tiny{\BV{USPS}$\rightarrow$\BV{MNIST} }} & 
 {\tiny{\BV{USPS}$\rightarrow$\BV{QMNIST} }} &    
{{{Avg}}} \\

\hline
{D2V}  & 28.2$\pm$0.31 & 94.5$\pm$0.63  & 27.6$\pm$0.41  & 29.3$\pm$0.39 & 44.9\\
D2V \textit{w/o. Gram} & 28.5$\pm$0.29 & 92.4$\pm$0.56  & 25.5$\pm$0.29  & 27.7$\pm$0.26 & 43.5\\
D2V \textit{w/o. Mutual} & 26.7$\pm$0.27 & 94.1$\pm$0.49  & 27.9$\pm$0.35  & 27.4$\pm$0.41 & 44.0\\
\end{tabular}

\begin{tabular}{c| c| c| c| c| c|| c|c|c|c||c }
{target} &
 {\tiny{VisDA}} &  
 {\tiny{Ytb BBox}} & 
  {\tiny{PASCAL}} & 
 {\tiny{COCO}} &  
 Avg&
  {\tiny{VisDA}} &  
 {\tiny{Ytb BBox}} & 
  {\tiny{PASCAL}} & 
 {\tiny{COCO}} &  
{{{Avg}}} \\

\hline
{D2V}  & 56.6 & 70.6  & 81.3 & 86.8 & 73.8 & 36.6 & 76.8 & 70.0& 78.8 & 65.5\\
D2V \textit{w/o. Gram} & 54.5 & 68.4  & 80.5  & 85.4 & 72.2 & 34.5 & 77.1 & 65.4 & 77.9 & 63.7\\
D2V \textit{w/o. Mutual} & 55.2 & 69.3 & 81.4 & 85.7 &72.9  &35.4 & 73.5 & 67.8 &77.5 & 63.5\\
\end{tabular}

} 
\caption{The ablation study results show that the Mutual information minimizing and Gram matrix information is essential for our model. The above table shows ablation experiments performed on the \TinyDA~dataset. The table below shows ablation experiments on \DomainBank~dataset (openset DA on the left, partial DA on the right). }
\label{table_ablation}

\end{table*}

\section{Conclusion}
In this paper, we have proposed a novel learning paradigm to explore the natural relations between different domains. We introduced the deep domain embedding task and proposed \ModelName~to achieve domain-to-vector mapping with joint learning of Gram Matrix of the latent representations and feature disentanglement. We have collected and evaluated two state-of-the-art domain adaptation datasets, \TinyDA~and \DomainBank. These two datasets are challenging due to the presence of notable domain gaps and a large number of domains. Extensive experiments has been conducted, both qualitatively and quantitatively, on the two benchmarks we collected to demonstrate the effectiveness of our proposed model. We also show that our model can facilitate multi-source domain adaptation, openset domain adaptation and partial domain adaptation.  We hope the learning schema we proposed and the benchmarks we collected will be beneficial for the future domain adaptation research.

\section*{Acknowledgements}
We thank the anonymous reviewers for their comments and suggestions. This work was partially supported by NSF and Honda Research Institute. 

\bibliographystyle{splncs04}
\bibliography{eccv_2020.bib}

\newpage
\vspace{-0.2cm}
\section{Supplementary Material}
\addcontentsline{toc}{section}{Appendices}
\renewcommand{\thesubsection}{\Alph{subsection}}
The appendix is organized as follows: Section~\ref{comparison_to_modern_dataset} shows the comparison of our two datasets with the state-of-the-art cross-domain datasets. Section~\ref{tinyda_generation} describes the details of generating the \TinyDA~dataset. Section~\ref{domainbank_dataset} shows the detailed information about \DomainBank~dataset. Section~\ref{model_arch} introduces the detailed network framework for experiments on \TinyDA~dataset. Section~\ref{additional_exp} shows the additional experimental analysis. Section~\ref{cate_gory} shows the category information in the openset domain adaptation experiments in Section~\ref{subsect_openset}.
\vspace{-0.3cm}
\subsection{Comparison to modern datasets}
\vspace{-0.8cm}
\label{comparison_to_modern_dataset}

\begin{table*}[h!]
{\fontfamily{<ptm>}\selectfont 
\small
\setlength{\tabcolsep}{0.2em}
    \centering
    \begin{tabular}{|c|c| c c c| c|}
    \hline
    Dataset & Year &\footnotesize{Images} & \footnotesize{Classes} & \footnotesize{Domains} & \textit{Description} \\
    \hline
    
    \footnotesize{Digit-Five}& - &$\sim$100,000 & 10 & 5 & digit \\
    \footnotesize{Office~\cite{office}} & 2010 & 4,110 & 31 & 3 & office\\
    \footnotesize{Office-Caltech~\cite{gong2012geodesic}}& 2012 & 2,533 & 10 & 4 & office \\
     \footnotesize{CAD-Pascal~\cite{peng2015learning}}& 2015 & 12,000 & 20 & 6 & \footnotesize{animal,vehicle} \\
    \footnotesize{Office-Home~\cite{officehome}} & 2017 & 15,500 & 65 & 4 
    & office, home\\
    \footnotesize{PACS}~\cite{PACS} & 2017 & 9,991 & 7 & 4 & animal, stuff\\
    \footnotesize{Open MIC~\cite{openmic}} & 2018 &  16,156 & - & - & museum\\
    \footnotesize{Syn2Real~\cite{syn2real}} & 2018 &  280,157 & 12 & 3 & \footnotesize{ animal,vehicle}\\ 
    
    DomainNet~\cite{domainnet} &2019& 569,010& 345 & 6 & clipart,sketch\\
    
    \hline
    \textbf{\TinyDA} (Ours)&-& \textbf{965,619} &\textbf{10 or 26}& \textbf{54}  &tiny images\\
    \textbf{DomainBank} (Ours)&-& \textbf{339,772}& \textbf{-} & \textbf{55} & \footnotesize{dataset}\\
    \hline
    \end{tabular}

    \caption{A collection of most notable datasets to evaluate domain adaptation methods. Specifically, ``Digit-Five'' dataset indicates five most popular digit datasets (\textit{MNIST}~\cite{lecun1998gradient}, \textit{MNIST-M}~\cite{DANN}, Synthetic Digits~\cite{DANN}, \textit{SVHN}, and \textit{USPS}) which are widely used to evaluate domain adaptation models.  Our dataset is challenging as it contains more images and domains than other datasets.}
    \label{tab_dataset}
}    
\end{table*}

\vspace{-1.5cm}

\subsection{\TinyDA~ Generation}
\label{tinyda_generation}
The images from \TinyDA~ dataset are generated by blending different foreground shapes over patches randomly extracted from background images. In the first step, we select a foreground shape from the following five MNIST-style datasets: MNIST~\cite{mnist}, USPS~\cite{usps}, EMNIST~\cite{emnist}, KMNIST~\cite{kmnist}, QMNIST~\cite{qmnist}, and  FashionMNIST~\cite{fashionmnist}. Secondly, we choose a background pattern from the CIFAR10~\cite{cifar10} dataset or randomly cropped from BSD500~\cite{bsds500} dataset. Thirdly, we perform three different post-process to our rendered images: (1) replace the background with black patch, (2) replace the background with white path, (3) convert the generated images to grayscale images. These three post-processes, together with the original foreground images and the generated color images, form a dataset with five different modes, \textit{i.e.} Black Background ({\BV{BB}}), White Background ({\BV{WB}}),  GrayScale image ({\BV{GS}}), Color ({\BV{Cr}}) image, Original image({\BV{Or}}). In total, we generate a dataset with 54 domains and about one million MNIST-style training examples. 

The image examples of our \TinyDA~dataset are shown in Table~\ref{tiny_da_image}. Specifically, the upper and below table show the images generated with backgrounds from BSDS500~\cite{bsds500} and CIFAR10~\cite{cifar10}, respectively. The image number of each domain in \TinyDA~dataset can be seen from Table~\ref{tab_tiny_da_statistical}.

\begin{figure*}[]
    \centering
    \includegraphics[width=0.8\linewidth]{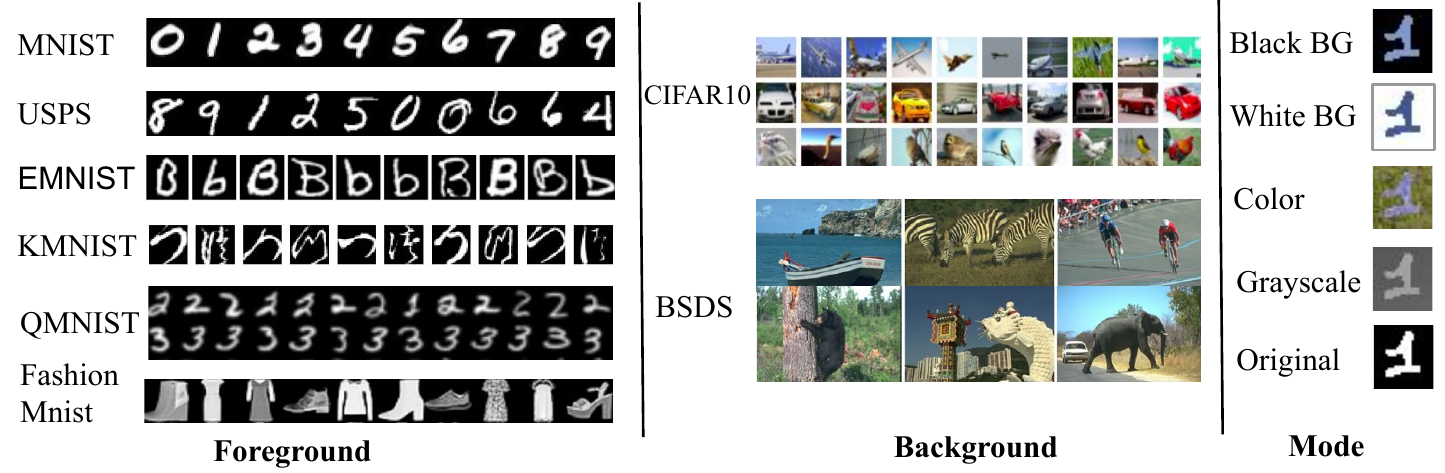}
    \caption{Generation configuration for \TinyDA~ dataset. We create our TinyDA dataset with six foregrounds, two backgrounds and five modes. The foreground images are from MNIST~\cite{mnist}, USPS~\cite{usps}, EMNIST~\cite{emnist}, KMNIST~\cite{kmnist}, QMNIST~\cite{qmnist}, FashionMNIST~\cite{fashionmnist}. The background images are randomly sampled from CIFAR10~\cite{cifar10} or randomly cropped from BSDS500~\cite{bsds500} dataset. The five modes include ``Black Background'', ``White Background'', ``Color'', ``GrayScale'', and ``Original''.}. 
    \label{fig_tiny_da_conf}
    \centering
    \vspace{-0.3cm}
\end{figure*}

\begin{table*}[t]
    \centering
    \begin{tabular}{c|c c c c c}
         FG/Mode& Black BG & White BG & Color & Grayscale & Original\\
         \Xhline{1.5pt}
         
         MNIST&\includegraphics[width=0.15\linewidth]{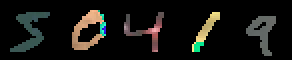} &\includegraphics[width=0.15\linewidth]{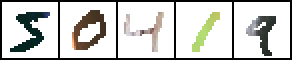} &\includegraphics[width=0.15\linewidth]{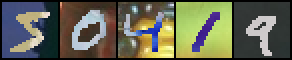} &\includegraphics[width=0.15\linewidth]{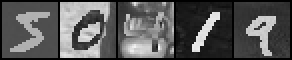} &\includegraphics[width=0.15\linewidth]{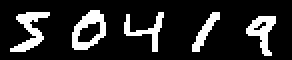}  \\
         
         USPS &\includegraphics[width=0.15\linewidth]{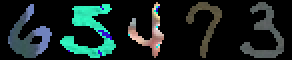} &\includegraphics[width=0.15\linewidth]{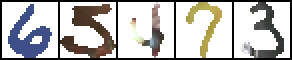} &\includegraphics[width=0.15\linewidth]{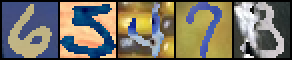} &\includegraphics[width=0.15\linewidth]{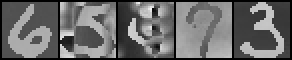} &\includegraphics[width=0.15\linewidth]{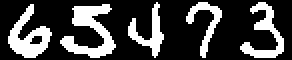} \\
         
FashionMNIST &\includegraphics[width=0.15\linewidth]{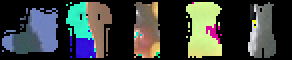} &\includegraphics[width=0.15\linewidth]{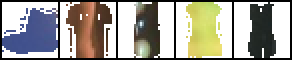} &\includegraphics[width=0.15\linewidth]{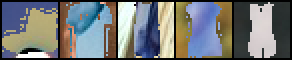} &\includegraphics[width=0.15\linewidth]{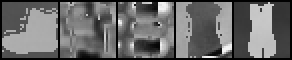} &\includegraphics[width=0.15\linewidth]{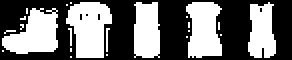} \\

KMNIST &\includegraphics[width=0.15\linewidth]{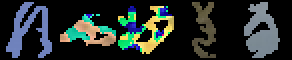} &\includegraphics[width=0.15\linewidth]{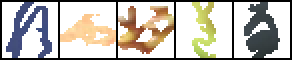} &\includegraphics[width=0.15\linewidth]{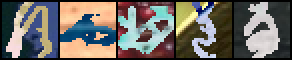} &\includegraphics[width=0.15\linewidth]{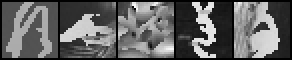} &\includegraphics[width=0.15\linewidth]{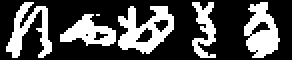} \\

QMNIST &\includegraphics[width=0.15\linewidth]{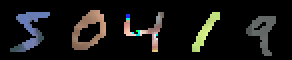} &\includegraphics[width=0.15\linewidth]{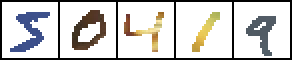} &\includegraphics[width=0.15\linewidth]{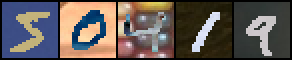} &\includegraphics[width=0.15\linewidth]{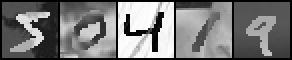} &\includegraphics[width=0.15\linewidth]{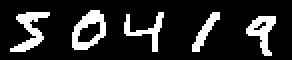} \\

EMNIST &\includegraphics[width=0.15\linewidth]{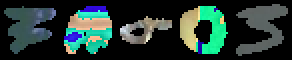} &\includegraphics[width=0.15\linewidth]{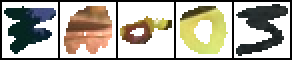} &\includegraphics[width=0.15\linewidth]{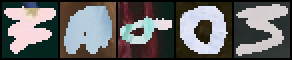} &\includegraphics[width=0.15\linewidth]{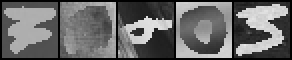} &\includegraphics[width=0.15\linewidth]{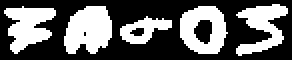} \\

\Xhline{1.2pt}

MNIST &\includegraphics[width=0.15\linewidth]{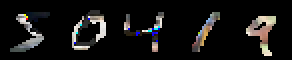} &\includegraphics[width=0.15\linewidth]{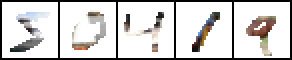} &\includegraphics[width=0.15\linewidth]{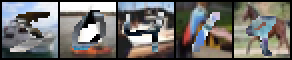} &\includegraphics[width=0.15\linewidth]{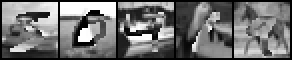} &\includegraphics[width=0.15\linewidth]{tmp/fg_mnist_bg_bsds500_model_original.png} \\

USPS &\includegraphics[width=0.15\linewidth]{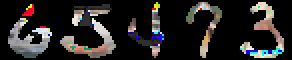} &\includegraphics[width=0.15\linewidth]{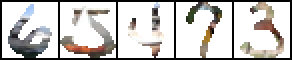} &\includegraphics[width=0.15\linewidth]{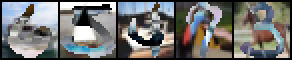} &\includegraphics[width=0.15\linewidth]{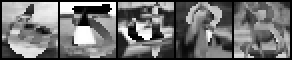} &\includegraphics[width=0.15\linewidth]{tmp/fg_usps_bg_bsds500_model_original.png} \\

FashionMNIST &\includegraphics[width=0.15\linewidth]{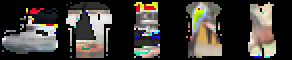} &\includegraphics[width=0.15\linewidth]{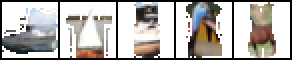} &\includegraphics[width=0.15\linewidth]{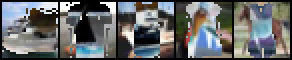} &\includegraphics[width=0.15\linewidth]{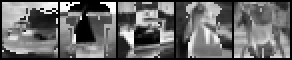} &\includegraphics[width=0.15\linewidth]{tmp/fg_fashionMnist_bg_bsds500_model_original.png} \\

KMNIST &\includegraphics[width=0.15\linewidth]{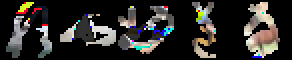} &\includegraphics[width=0.15\linewidth]{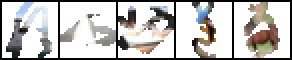} &\includegraphics[width=0.15\linewidth]{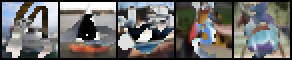} &\includegraphics[width=0.15\linewidth]{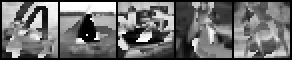} &\includegraphics[width=0.15\linewidth]{tmp/fg_kMnist_bg_bsds500_model_original.png} \\

QMNIST &\includegraphics[width=0.15\linewidth]{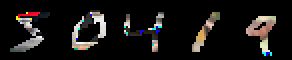} &\includegraphics[width=0.15\linewidth]{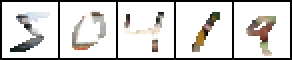} &\includegraphics[width=0.15\linewidth]{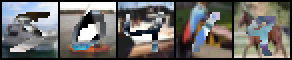} &\includegraphics[width=0.15\linewidth]{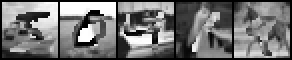} &\includegraphics[width=0.15\linewidth]{tmp/fg_qMnist_bg_bsds500_model_original.png} \\

EMNIST &\includegraphics[width=0.15\linewidth]{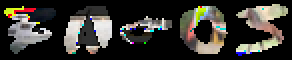} &\includegraphics[width=0.15\linewidth]{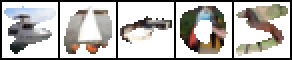} &\includegraphics[width=0.15\linewidth]{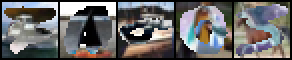} &\includegraphics[width=0.15\linewidth]{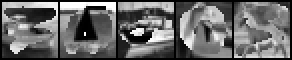} &\includegraphics[width=0.15\linewidth]{tmp/fg_eMnist_bg_bsds500_model_original.png} \\

    \end{tabular}
    \caption{Illustration of \TinyDA~dataset. We create our \TinyDA~dataset with six foregrounds, two backgrounds, and five modes. The upper and below table show the images generated with backgrounds from BSDS500~\cite{bsds500} and CIFAR10~\cite{cifar10}, respectively. } 
    \label{tiny_da_image}
\end{table*}
\begin{table}[]
    \centering
    \begin{tabular}{c|c | c | p{1.4cm} |c| c}
         FG/Mode& Black BG & White BG & Color & Grayscale & Original\\
         \Xhline{0.7pt}
         MNIST & 40,000& 40,000& 40,000&40,000 & 20,000 \\
         USPS &14,582 & 14,582& 14,582& 14,582& 7291\\
         FashionMNIST & 40,000& 40,000& 40,000&40,000 & 20,000 \\
         KMNIST & 40,000& 40,000& 40,000&40,000 & 20,000\\
         QMNIST & 40,000& 40,000& 40,000&40,000 & 20,000 \\
         EMNIST & 40,000& 40,000& 40,000&40,000 & 20,000 \\
    \end{tabular}
    \caption{Number of images in each domain of \TinyDA~dataset.}
    \label{tab_tiny_da_statistical}
\end{table}

\clearpage
\begin{table*}[t]
    \centering
    \begin{tabular}{c|c |c |c }
         ID & Image Samples & ID & Image Samples\\
         \Xhline{0.7pt}
   1&\includegraphics[width=0.40\linewidth]{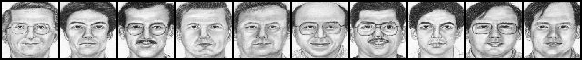}&
2&\includegraphics[width=0.40\linewidth]{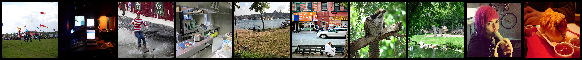}\\
3&\includegraphics[width=0.40\linewidth]{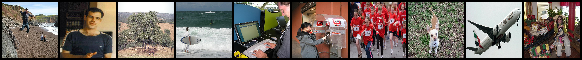}&
4&\includegraphics[width=0.40\linewidth]{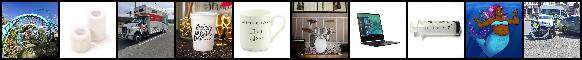}\\
5&\includegraphics[width=0.40\linewidth]{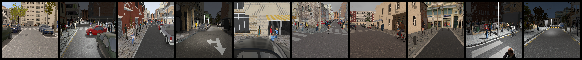}&
6&\includegraphics[width=0.40\linewidth]{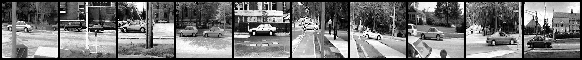}\\
7&\includegraphics[width=0.40\linewidth]{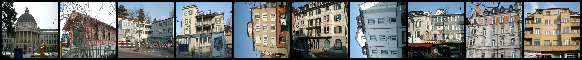}&
8&\includegraphics[width=0.40\linewidth]{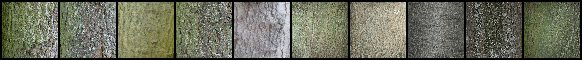}\\
9&\includegraphics[width=0.40\linewidth]{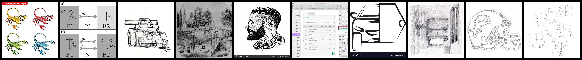}&
10&\includegraphics[width=0.40\linewidth]{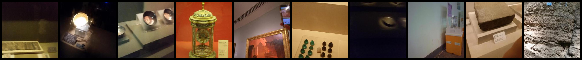}\\
11&\includegraphics[width=0.40\linewidth]{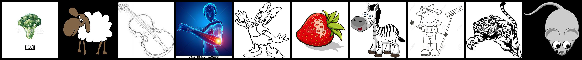}&
12&\includegraphics[width=0.40\linewidth]{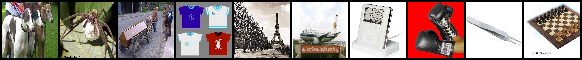}\\
13&\includegraphics[width=0.40\linewidth]{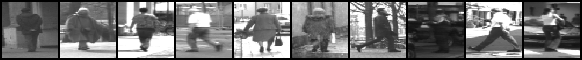}&
14&\includegraphics[width=0.40\linewidth]{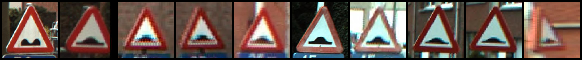}\\
15&\includegraphics[width=0.40\linewidth]{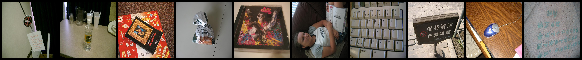}&
16&\includegraphics[width=0.40\linewidth]{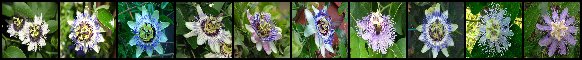}\\
17&\includegraphics[width=0.40\linewidth]{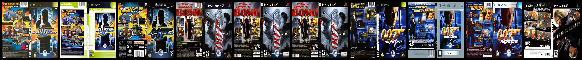}&
18&\includegraphics[width=0.40\linewidth]{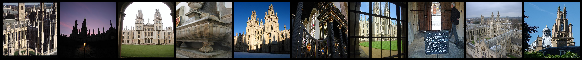}\\
19&\includegraphics[width=0.40\linewidth]{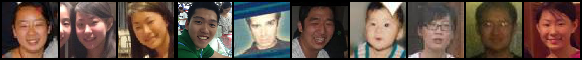}&
20&\includegraphics[width=0.40\linewidth]{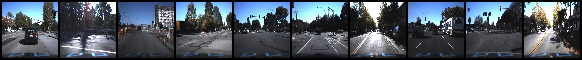}\\
21&\includegraphics[width=0.40\linewidth]{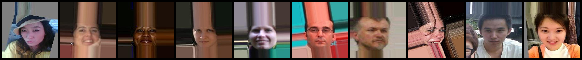}&
22&\includegraphics[width=0.40\linewidth]{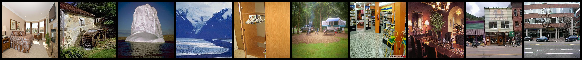}\\
23&\includegraphics[width=0.40\linewidth]{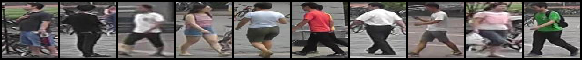}&
24&\includegraphics[width=0.40\linewidth]{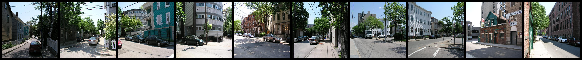}\\
25&\includegraphics[width=0.40\linewidth]{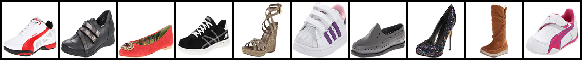}&
26&\includegraphics[width=0.40\linewidth]{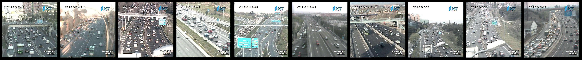}\\
27&\includegraphics[width=0.40\linewidth]{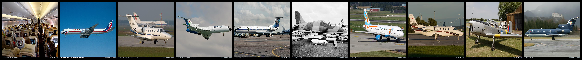}&
28&\includegraphics[width=0.40\linewidth]{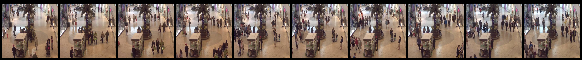}\\
29&\includegraphics[width=0.40\linewidth]{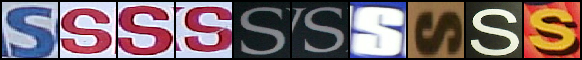}&
30&\includegraphics[width=0.40\linewidth]{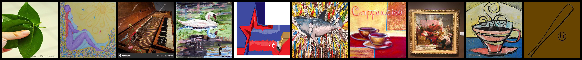}\\
31&\includegraphics[width=0.40\linewidth]{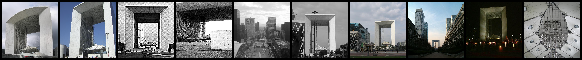}&
32&\includegraphics[width=0.40\linewidth]{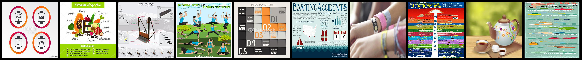}\\
33&\includegraphics[width=0.40\linewidth]{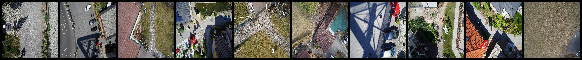}&
34&\includegraphics[width=0.40\linewidth]{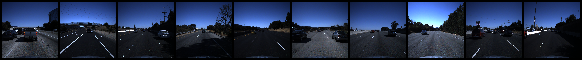}\\
35&\includegraphics[width=0.40\linewidth]{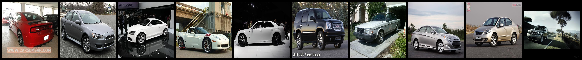}&
36&\includegraphics[width=0.40\linewidth]{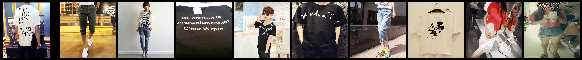}\\
37&\includegraphics[width=0.40\linewidth]{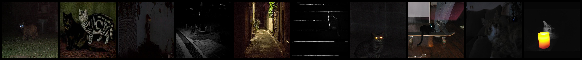}&
38&\includegraphics[width=0.40\linewidth]{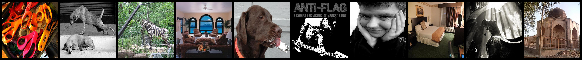}\\
39&\includegraphics[width=0.40\linewidth]{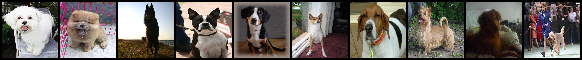}&
40&\includegraphics[width=0.40\linewidth]{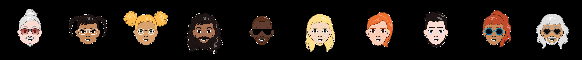}\\
41&\includegraphics[width=0.40\linewidth]{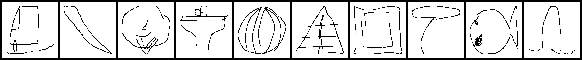}&
42&\includegraphics[width=0.40\linewidth]{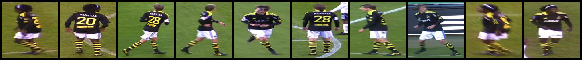}\\
43&\includegraphics[width=0.40\linewidth]{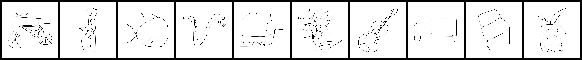}&
44&\includegraphics[width=0.40\linewidth]{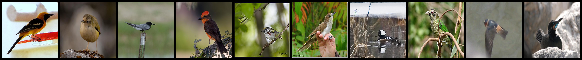}\\
45&\includegraphics[width=0.40\linewidth]{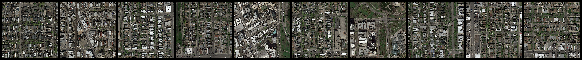}&
46&\includegraphics[width=0.40\linewidth]{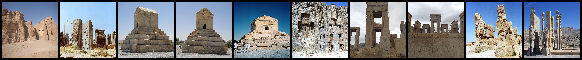}\\
47&\includegraphics[width=0.40\linewidth]{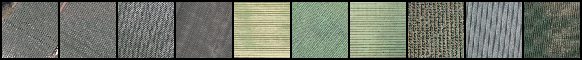}&
48&\includegraphics[width=0.40\linewidth]{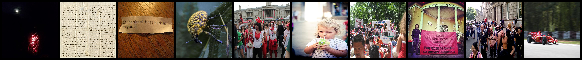}\\
49&\includegraphics[width=0.40\linewidth]{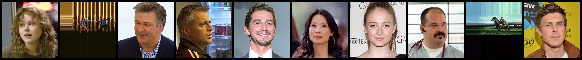}&
50&\includegraphics[width=0.40\linewidth]{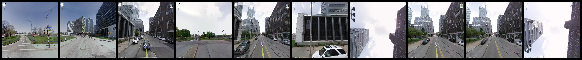}\\
51&\includegraphics[width=0.40\linewidth]{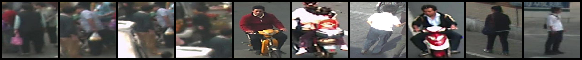}&
52&\includegraphics[width=0.40\linewidth]{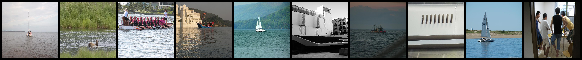}\\
53&\includegraphics[width=0.40\linewidth]{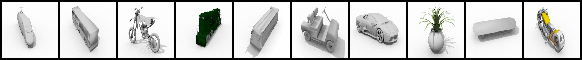}&
54&\includegraphics[width=0.40\linewidth]{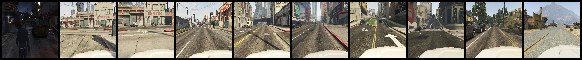}\\
55&\includegraphics[width=0.40\linewidth]{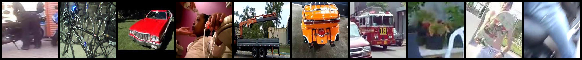}&
56&\includegraphics[width=0.40\linewidth]{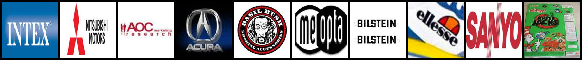}\\
    \end{tabular}
    \caption{Illustration of \DomainBank~dataset. The ID is this table is corresponding to the ID in Table~\ref{domain_bank_link}.} 
    \label{figure_domainbank_image}
\end{table*}
\clearpage
\subsection{\DomainBank~ Dataset}
\label{domainbank_dataset}
The images of~\DomainBank~ dataset are sampled from 56 existing popular computer vision datasets. We choose the dataset with different image modalities, illuminations, camera perspectives \textit{etc.} to increase the diversity of the domains. More details about our \DomainBank~benchmark are shown in Table~\ref{figure_domainbank_image} and Table~\ref{domain_bank_link}. In total, we collect 339,772 images with image-level and domain-level annotations. Different from \TinyDA, the categories of different domains in \DomainBank~ are not identical. This property makes \DomainBank~a good testbed for Openset Domain Adaptation and Partial Domain Adaptation.

\newcounter{Counter}

\begin{table*}[t]
{\fontfamily{<ptm>}\selectfont 
\scriptsize

    \centering
    \begin{tabular}{|c|c| c|c|| c|c| c|c|}
    \hline
    ID & Dataset Name & Image\# & Description& ID & Dataset Name & Image\# & Description \\
    \hline
    \addtocounter{Counter}{1}\arabic{Counter} &\href{http://mmlab.ie.cuhk.edu.hk/archive/cufsf/}{\color{blue}CUFSF}
    ~\cite{ucfsf} & 1,194 & face\_sketch&\addtocounter{Counter}{1}\arabic{Counter} & \href{http://cocodataset.org/#home}{\color{blue}COCO}
    ~\cite{mscoco} & 10,000 & real\\
    
    \addtocounter{Counter}{1}\arabic{Counter} &\href{http://host.robots.ox.ac.uk/pascal/VOC/}{\color{blue}PASCAL}
    ~\cite{pascal} & 10,000&realh&\addtocounter{Counter}{1}\arabic{Counter} & \href{http://ai.bu.edu/M3SDA/}{\color{blue}DomainNet}
    ~\cite{domainnet} & 10,000 & real\\
    
    \addtocounter{Counter}{1}\arabic{Counter} &\href{  http://synthia-dataset.net/}{\color{blue}SYNTHIA}
    ~\cite{synthia} & 10,000& street\_syn &
    \addtocounter{Counter}{1}\arabic{Counter} & \href{  http://l2r.cs.uiuc.edu/~cogcomp/Data/Car/}{\color{blue}UIUC CAR}
    ~\cite{UIUC_CAR} & 1,220 & car\\
    
    \addtocounter{Counter}{1}\arabic{Counter} &\href{
    http://www.vision.ee.ethz.ch/showroom/zubud/index.de.html
    }{\color{blue}ZuDuB}
    ~\cite{ZuBuD} & 210&building&\addtocounter{Counter}{1}\arabic{Counter} & \href{
    http://eidolon.univ-lyon2.fr/~remi1/Bark-101/
    }{\color{blue}Bark-101}
    ~\cite{bark} & 2,586 & bark\\

    \addtocounter{Counter}{1}\arabic{Counter} &\href{
    http://ai.bu.edu/M3SDA/
    }{\color{blue}DomainNet}
    ~\cite{domainnet} & 10,000&sketch&\addtocounter{Counter}{1}\arabic{Counter} & \href{
    http://users.cecs.anu.edu.au/~koniusz/openmic-dataset/
    }{\color{blue}Open-MIC}
    ~\cite{openmic} & 10,000 & indoor\\
    
    \addtocounter{Counter}{1}\arabic{Counter} &\href{
    http://ai.bu.edu/M3SDA/}{\color{blue}DomainNet}
    ~\cite{domainnet} & 10,000&clipart&\addtocounter{Counter}{1}\arabic{Counter} & \href{
    http://www.vision.caltech.edu/Image_Datasets/Caltech256/}{\color{blue}Caltech256}
    ~\cite{griffin2007caltech} & 10,000 & real\\
    
    \addtocounter{Counter}{1}\arabic{Counter} &\href{http://www.gavrila.net/Datasets/Daimler_Pedestrian_Benchmark_D/Daimler_Mono_Ped__Detection_Be/daimler_mono_ped__detection_be.html}{\color{blue}Ped. Detection}
    ~\cite{enzweiler2008monocular} & 10,000&pedestrian&\addtocounter{Counter}{1}\arabic{Counter} & \href{ http://www.vision.ee.ethz.ch/~timofter/traffic_signs/}{\color{blue}Traffic Sign}
    ~\cite{mathias2013traffic} & 4,053& traffic\\
    
    \addtocounter{Counter}{1}\arabic{Counter} &\href{https://archive.org/details/ukbench }{\color{blue}UKBench}
    ~\cite{nister2006scalable} & 10,000&indoor\_stuff&\addtocounter{Counter}{1}\arabic{Counter} & \href{ https://www.robots.ox.ac.uk/~vgg/data/flowers/ }{\color{blue}Oxford Flower}
    ~\cite{nilsback2006visual} & 8,189 & flower\\
    
    \addtocounter{Counter}{1}\arabic{Counter} &\href{http://www.vision.caltech.edu/malaa/datasets/caltech-games/}{\color{blue}Caltech Games}
    ~\cite{aly2009towards} & 7,660&game\_cover&\addtocounter{Counter}{1}\arabic{Counter} & \href{http://www.robots.ox.ac.uk/~vgg/data/oxbuildings/}{\color{blue}Oxford Buildings}
    ~\cite{philbin2007object} & 5,063 & building\\
    
    \addtocounter{Counter}{1}\arabic{Counter} &\href{https://github.com/bj80heyue/Learning-to-Group}{\color{blue}GFW Face}
    ~\cite{he2018merge} & 3,236&face&\addtocounter{Counter}{1}\arabic{Counter} & \href{ https://github.com/udacity/self-driving-car/tree/master/annotations}{\color{blue}Driving}
    ~\cite{udacity} & 9,420 & road\\
    
    \addtocounter{Counter}{1}\arabic{Counter} &\href{http://mmlab.ie.cuhk.edu.hk/projects/MegaAge/}{\color{blue}MegaAge}
    ~\cite{huang2016unsupervised} & 10,000&face&\addtocounter{Counter}{1}\arabic{Counter} & \href{ http://groups.csail.mit.edu/vision/datasets/ADE20K/}{\color{blue}ADE20K}
    ~\cite{zhou2019semantic} & 10,000 & indoor\\
    
    \addtocounter{Counter}{1}\arabic{Counter} &\href{ http://mmlab.ie.cuhk.edu.hk/projects/PCN.html}{\color{blue}Ped. Color}
    ~\cite{cheng2016pedestrian} & 10,000&pedestrian&\addtocounter{Counter}{1}\arabic{Counter} & \href{http://www.inf-cv.uni-jena.de/Group/Staff/Former+Members/Dr_+Bj\%C3\%B6rn+Fr\%C3\%B6hlich-p-54/LabelMeFacade+Database.html }{\color{blue}LabelMeFacade}
    ~\cite{labelmefacade} & 395 & building\\

    \addtocounter{Counter}{1}\arabic{Counter} &\href{ http://vision.cs.utexas.edu/projects/finegrained/utzap50k/ }{\color{blue}UT Zappos50K
}
    ~\cite{finegrained} & 10,000&shoes&\addtocounter{Counter}{1}\arabic{Counter} & \href{ http://agamenon.tsc.uah.es/Personales/rlopez/data/trancos/}{\color{blue}TRANCOS}
    ~\cite{TRANCOSdataset_IbPRIA2015} & 1,244 & traffic\\
    
    \addtocounter{Counter}{1}\arabic{Counter} &\href{http://www.robots.ox.ac.uk/~vgg/data/fgvc-aircraft/ }{\color{blue}FGVC}
    ~\cite{TRANCOSdataset_IbPRIA2015} & 10,000&aeroplane&\addtocounter{Counter}{1}\arabic{Counter} & \href{ http://personal.ie.cuhk.edu.hk/~ccloy/downloads_mall_dataset.html}{\color{blue}Mall Dataset}
    ~\cite{change2013semi} & 2,000 & mall\\
    
    \addtocounter{Counter}{1}\arabic{Counter} &\href{ http://www.ee.surrey.ac.uk/CVSSP/demos/chars74k/}{\color{blue}Chars74K}
    ~\cite{de2009character} & 7,705&character&\addtocounter{Counter}{1}\arabic{Counter} & \href{ http://ai.bu.edu/M3SDA/}{\color{blue}DomainNet}
    ~\cite{domainnet} &10,000 & painting\\
    
    \addtocounter{Counter}{1}\arabic{Counter} &\href{http://www.robots.ox.ac.uk/~vgg/data/parisbuildings/}{\color{blue}Paris Dataset}
    ~\cite{philbin2008lost} & 3,187&street&\addtocounter{Counter}{1}\arabic{Counter} & \href{http://ai.bu.edu/M3SDA/}{\color{blue}DomainNet}
    ~\cite{domainnet} & 10,000 & infograph\\

    \addtocounter{Counter}{1}\arabic{Counter} &\href{http://dronedataset.icg.tugraz.at/}{\color{blue}DroneDataset}
    ~\cite{dronedataset} & 400&drone &\addtocounter{Counter}{1}\arabic{Counter} & \href{https://boxy-dataset.com/}{\color{blue}Boxy}
    ~\cite{boxy2019} & 2,148 & road\\
    
    \addtocounter{Counter}{1}\arabic{Counter} &\href{https://ai.stanford.edu/~jkrause/cars/car_dataset.html}{\color{blue}Stanford Car}
    ~\cite{stanfordcar} & 8,144&car&\addtocounter{Counter}{1}\arabic{Counter} & \href{ https://github.com/switchablenorms/DeepFashion2}{\color{blue}DeepFashion2}
    ~\cite{DeepFashion2} & 10,000 & fashion\\
    
    \addtocounter{Counter}{1}\arabic{Counter} &\href{ https://github.com/cs-chan/Exclusively-Dark-Image-Dataset}{\color{blue}ExDark}
    ~\cite{Exdark} & 6,619&dark&\addtocounter{Counter}{1}\arabic{Counter} & \href{ http://memorability.csail.mit.edu/}{\color{blue}LaMem}
    ~\cite{ICCV15_Khosla} & 10,000 & memorial\\
    
    \addtocounter{Counter}{1}\arabic{Counter} &\href{http://vision.stanford.edu/aditya86/ImageNetDogs/}{\color{blue}Stanford Dog}
    ~\cite{KhoslaYaoJayadevaprakashFeiFei_FGVC2011} & 10,000&dog&\addtocounter{Counter}{1}\arabic{Counter} & \href{https://google.github.io/cartoonset/}{\color{blue}Cartoon Set}
    ~\cite{royer2020xgan} & 9,999 & cartoon\\

    \addtocounter{Counter}{1}\arabic{Counter} &\href{http://ai.bu.edu/M3SDA/}{\color{blue}DomainNet}
    ~\cite{domainnet} & 10,000&quick\_draw&\addtocounter{Counter}{1}\arabic{Counter} & \href{http://www.csc.kth.se/~vahidk/football_data.html}{\color{blue}Football}
    ~\cite{kazemi2012using} & 771 & football\\
    
    \addtocounter{Counter}{1}\arabic{Counter} &\href{ http://cybertron.cg.tu-berlin.de/eitz/projects/classifysketch/}{\color{blue}Sketch Objects}
    ~\cite{eitz2012hdhso} & 10,000&sketch&\addtocounter{Counter}{1}\arabic{Counter} & \href{ http://www.vision.caltech.edu/visipedia/CUB-200.html}{\color{blue}CUB200}
    ~\cite{welinder2010caltech} & 10,000 & bird\\
    
    \addtocounter{Counter}{1}\arabic{Counter} &\href{ https://zenodo.org/record/1154821#.Wmb4MFQ-duW}{\color{blue}CITY-OSM}
    ~\cite{kaiser2017learning} & 914&drone\_view&\addtocounter{Counter}{1}\arabic{Counter} & \href{https://sites.google.com/site/zhexuutssjtu/projects/arch}{\color{blue}Arch Style}
    ~\cite{xu2014architectural} & 4,630 & building\\
    
    \addtocounter{Counter}{1}\arabic{Counter} &\href{http://weegee.vision.ucmerced.edu/datasets/landuse.html}{\color{blue}UCM Land}
    ~\cite{yang2010bag} & 2,100&satellite&\addtocounter{Counter}{1}\arabic{Counter} & \href{https://tribhuvanesh.github.io/vpa/}{\color{blue}Privacy Attribute}
    ~\cite{orekondy17iccv} &  4,157& stuff\\
    
        \addtocounter{Counter}{1}\arabic{Counter} &\href{https://data.vision.ee.ethz.ch/cvl/rrothe/imdb-wiki/}{\color{blue}IMDB-WIKI}
    ~\cite{Rothe-IJCV-2018} & 10,000&face&\addtocounter{Counter}{1}\arabic{Counter} & \href{http://crcv.ucf.edu/projects/GMCP_Geolocalization/}{\color{blue}Street View}
    ~\cite{6710175} & 6,594 & street\\

    \addtocounter{Counter}{1}\arabic{Counter} &\href{http://mmlab.ie.cuhk.edu.hk/projects/luoWTiccv2013DDN/index.html}{\color{blue}PPSS}
    ~\cite{luo2013pedestrian} & 1,458&pedestrian &\addtocounter{Counter}{1}\arabic{Counter} & \href{http://cybertron.cg.tu-berlin.de/eitz/tvcg_benchmark/index.html}{\color{blue}Sketch Retrieval}
    ~\cite{eitz2011sbir} & 1,213 & sketch\\
    
    \addtocounter{Counter}{1}\arabic{Counter} &\href{https://github.com/VisionLearningGroup/taskcv-2017-public/tree/master/classification}{\color{blue}VisDA}
    ~\cite{peng2017visda} & 10,000&syn&\addtocounter{Counter}{1}\arabic{Counter} & \href{https://download.visinf.tu-darmstadt.de/data/from_games/}{\color{blue}GTA}
    ~\cite{Richter_2016_ECCV} & 5,000 & syn\\
    
    \addtocounter{Counter}{1}\arabic{Counter} &\href{https://research.google.com/youtube-bb/}{\color{blue}Youtube BBox}
    ~\cite{real2017youtube} & 10,000&real&\addtocounter{Counter}{1}\arabic{Counter} & \href{https://github.com/msn199959/Logo-2k-plus-Dataset}{\color{blue}Logo-2k+}
    ~\cite{wang2019logo} & 10,000 & logo\\

 \hline

    \end{tabular}
    \caption{Detailed information about our \DomainBank~dataset.}
    \label{domain_bank_link}
}    
\end{table*}

\subsection{Model architecture}
\label{model_arch}
The detailed network architecture for \TinyDA~dataset is shown in Table~\ref{tab:digit_arch}.
\begin{table}[!h]
    \centering
    \begin{tabular}{c|l}
        \noalign{\hrule height 1pt}
        layer & configuration \\
        \noalign{\hrule height 1pt}
        \multicolumn{2}{c}{Feature Generator} \\
        \noalign{\hrule height 1pt}
        1 & Conv2D (3, 64, 5, 1, 2), BN, ReLU, MaxPool \\
        \hline
        2 & Conv2D (64, 64, 5, 1, 2), BN, ReLU, MaxPool \\
        \hline
        3 & Conv2D (64, 128, 5, 1, 2), BN, ReLU \\
        \noalign{\hrule height 1pt}     
        \multicolumn{2}{c}{Disentangler} \\
        \noalign{\hrule height 1pt}
        1 & FC (8192, 3072), BN, ReLU\\
        \hline
        2 &  DropOut (0.5), FC (3072, 2048), BN, ReLU \\
        \noalign{\hrule height 1pt}     
        \multicolumn{2}{c}{Domain Classifier} \\
        \noalign{\hrule height 1pt}
        1 & FC (2048, 256), LeakyReLU \\
        \hline
        2 & FC (256, 56), LeakyReLU \\
        \noalign{\hrule height 1pt}     
        \multicolumn{2}{c}{Classifier} \\
        \noalign{\hrule height 1pt}
        1 & FC (2048, 10 or 26), BN, Softmax \\
        \noalign{\hrule height 1pt}     
        \multicolumn{2}{c}{Reconstructor} \\
        \noalign{\hrule height 1pt}
        1 & FC (4096, 8192) \\
        \noalign{\hrule height 1pt}     
        \multicolumn{2}{c}{Mutual Information Estimator} \\
        \noalign{\hrule height 1pt}
        fc1\_x & FC (2048, 512) \\
        \hline
        fc1\_y & FC (2048, 512), LeakyReLU \\
        \hline
        2 & FC (512,1)\\
        \noalign{\hrule height 1pt}
    \end{tabular}
     \caption{Model Architecture for experiments on \TinyDA~ dataset. For each convolution layer, we list the input dimension, output dimension, kernel size, stride, and padding. For the fully-connected layer, we provide the input and output dimensions. For drop-out layers, we provide the probability of an element to be zeroed.}
        \label{tab:digit_arch}
\end{table}

\clearpage
\subsection{Additional experimental results}
\label{additional_exp}

\begin{figure*}[h!]
    \begin{minipage}{\hsize}
      \centering
      \subfigure[\scriptsize t-SNE Plot by BG ]
      {\includegraphics[width=0.46\hsize]{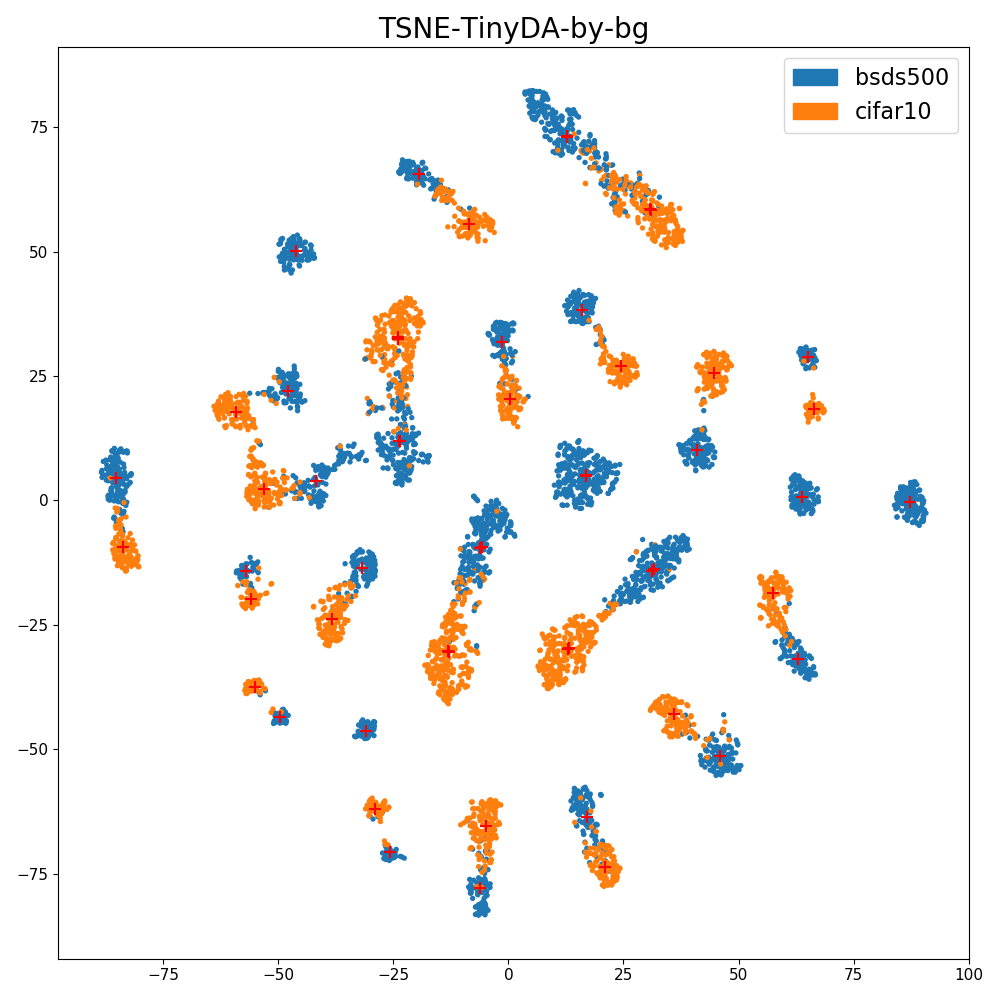}
      \label{tsne_tiny_da_bg} }
     \centering
      \subfigure[\scriptsize t-SNE Plot by FG]
      {\includegraphics[width=0.46\hsize]{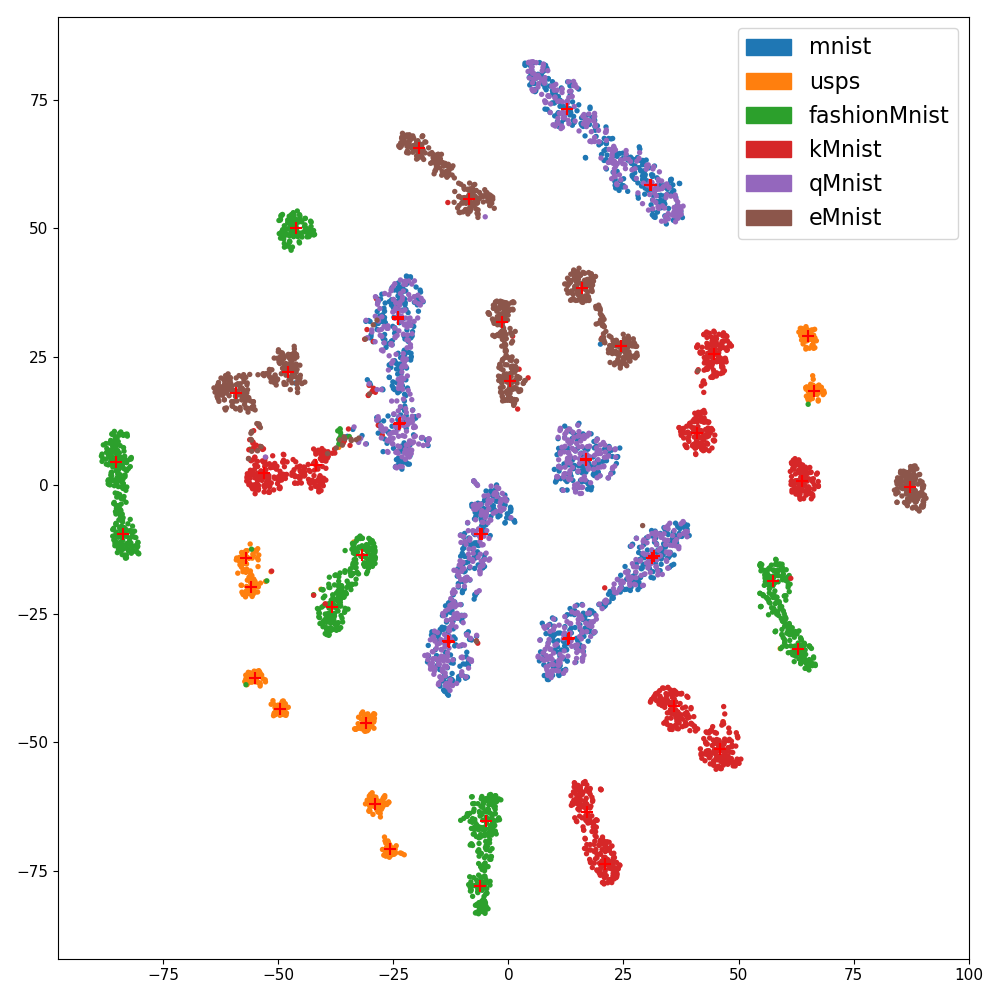}
      \label{tsne_tiny_da_fg}}
    \centering
      \subfigure[\scriptsize t-SNE Plot by Mode]
      {\includegraphics[width=0.46\hsize]{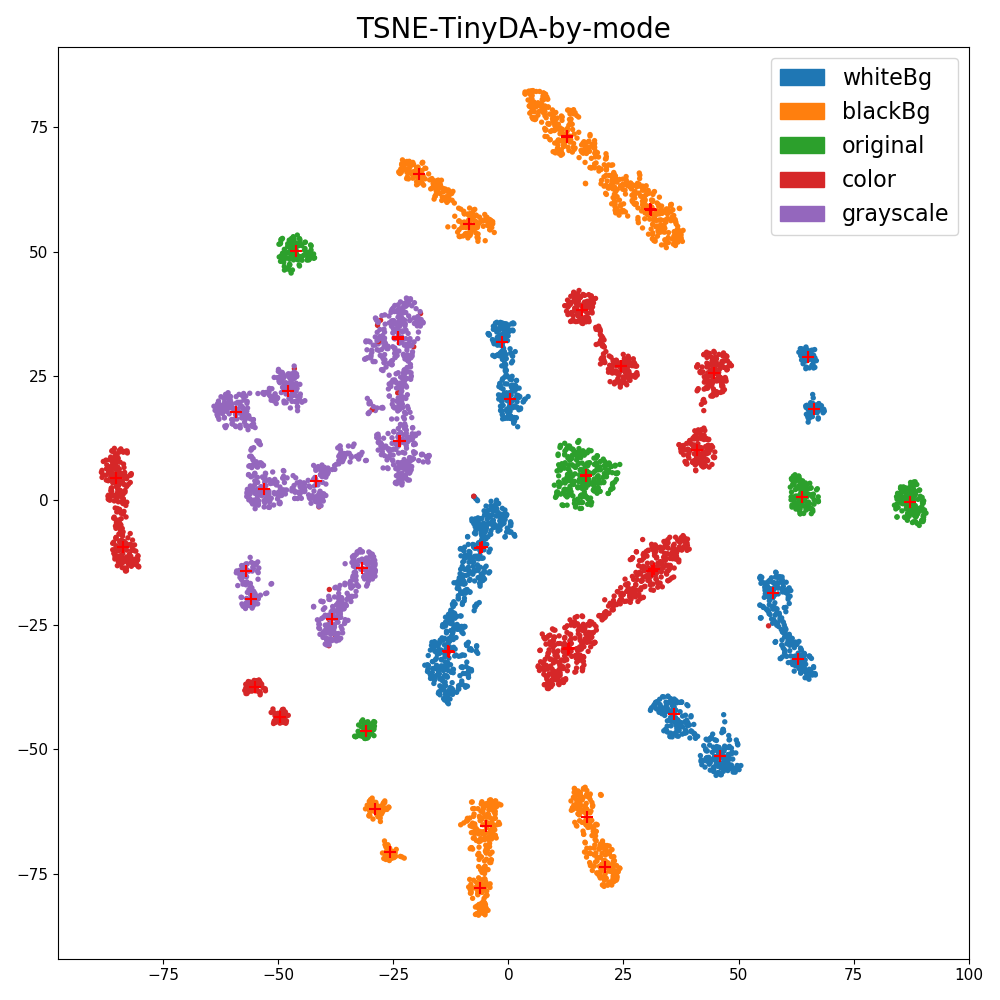}
      \label{tsne_tiny_da_mode} }
          \centering
      \subfigure[\scriptsize Deep Embedding]
      {\includegraphics[width=0.46\hsize]{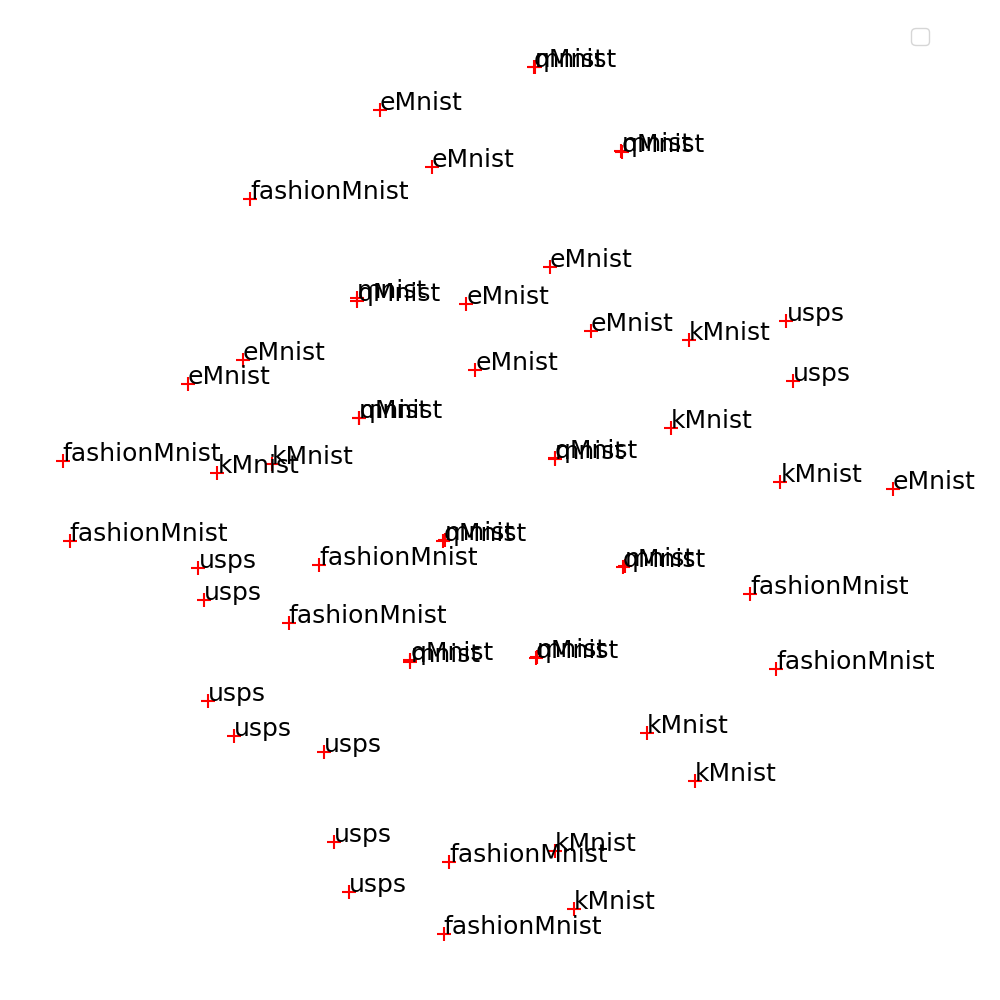}
      \label{tsne_tiny_da_embedding_supp}}
    \end{minipage}
  \caption{\footnotesize Deep domain embedding results of our \ModelName~model on \TinyDA~dataset: (\textbf{a}) t-SNE plot of the embedding result (color indicates different background); (\textbf{b})t-SNE plot of the embedding result (color indicates different foreground); (\textbf{c}) t-SNE plot of the embedding result (color indicates different mode); (\textbf{d}) Deep embedding result. (Best viewed in color. Zoom in to see details.)}
  \label{domain_embedding_tinyda_supp}
\end{figure*}

\begin{figure*}[h!]
    \begin{minipage}{\hsize}
      \centering
      \subfigure[\scriptsize t-SNE Plot by Domain]
      {\includegraphics[width=0.46\hsize]{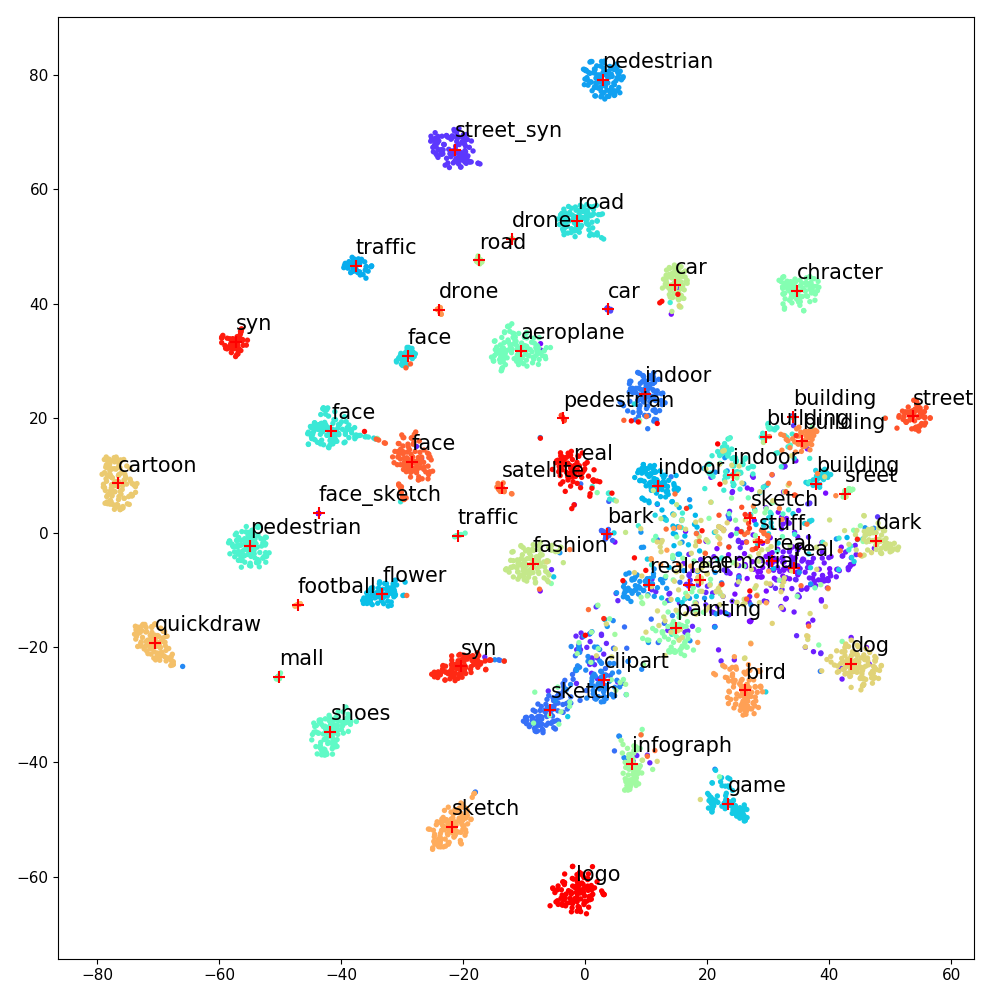}
      \label{tsne_domainbank_bg} }
          \centering
      \subfigure[\scriptsize Deep Embedding of DomainBank]
      {\includegraphics[width=0.46\hsize]{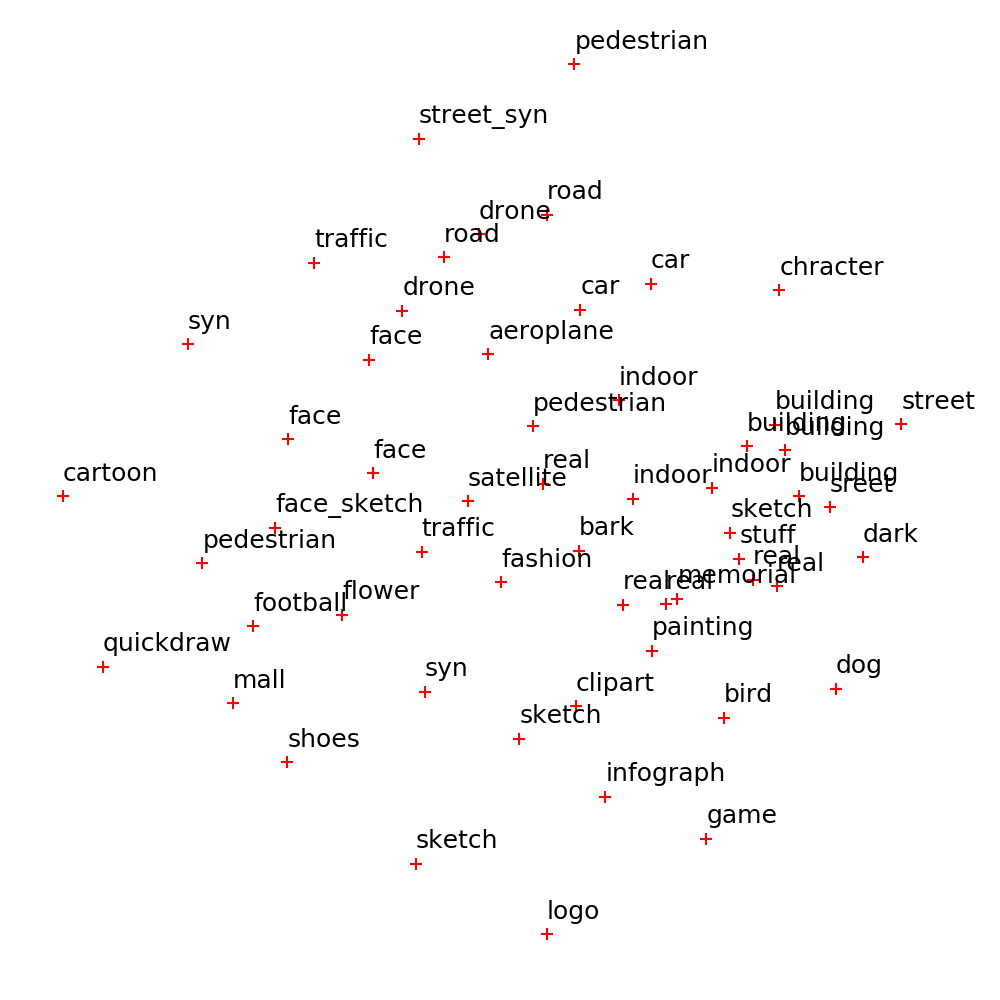}
      \label{tsne_tdomainbank_embedding_supp} }
    \end{minipage}
  \caption{\footnotesize Deep domain embedding results of our \ModelName~model on \DomainBank~dataset: (\textbf{a}) t-SNE plot of the embedding result (color indicates different domain); (\textbf{d}) Deep embedding result. (Best viewed in color. Zoom in to see details.)}
  \label{domain_embedding_domainnet_supp}
\end{figure*}

\subsection{Category information}
\label{cate_gory}
For openset domain adaptation experiments in Section~\ref{subsect_openset}, we choose the ``aeroplane'', ``bus'', ``horse'', ``motorcycle'', ``plant'', ``train'', and ``truck'' as the common categories across the four domains. We set ``bicycle'', ``car''  ``knife'', ``person'', ``skateboard'' as the unknown categories.





\end{document}